\RequirePackage{fix-cm}
\documentclass[smallextended]{svjour3}       
\smartqed  
\usepackage{graphicx}
\usepackage{amsmath,amsfonts,mathrsfs,amssymb}
\usepackage{bm}
\usepackage{algorithm}
\usepackage[noend]{algorithmic}
\usepackage{hyperref}
\usepackage{subfig}
\usepackage{color}
\usepackage{rotating}
\usepackage{multirow}
\usepackage{placeins}
\usepackage{soul}
\newcommand{\pluseq}{\mathrel{+}=}

\newcommand{\var}{\textit{Var}}

\newcommand{\dataset}{\mathscr{D}}
\newcommand{\datatrain}{\dataset_\text{TRAIN}}
\newcommand{\datatest}{\dataset_\text{TEST}}
\newcommand{\forest}{\mathcal{E}}
\newcommand{\tree}{\mathcal{T}}
\newcommand{\node}{\mathscr{N}}
\newcommand{\importance}{\mathit{importance}}
\newcommand{\error}{\mathit{e}}
\newcommand{\impurity}{\mathit{impu}}
\newcommand{\prototype}{\mathit{prototype}}
\newcommand{\gini}{\mathit{Gini}}

\newcommand{\yDomain}{\mathcal{Y}}

\newcommand{\labelset}{\mathscr{L}}

\newcommand{\binaryProblemExamples}[1]{\mathit{#1}}
\newcommand{\tp}{\binaryProblemExamples{tp}}
\newcommand{\fp}{\binaryProblemExamples{fp}}
\newcommand{\fn}{\binaryProblemExamples{fn}}
\newcommand{\pooled}{$\operatorname{AU}\overline{\mbox{PRC}}$}

\newcommand{\STAB}[1]{\begin{tabular}{@{}c@{}}#1\end{tabular}}

\newtheorem{assumption}{Assumption}

\journalname{Machine Learning Journal}
\begin{document}

\title{Feature Ranking for Semi-supervised Learning\thanks{We acknowledge the financial support of the Slovenian Research Agency via the grant P2-0103 and a young researcher grant to MP. SD and DK also acknowledge the support by the Slovenian Research Agency (via grants J7-9400, J7-1815, J2-9230, and N2-0128), and the European Commission (project AI4EU).}
}


\author{Matej Petkovi\'{c} \and
        Sa\v{s}o D\v{z}eroski \and
        Dragi Kocev
}


\institute{M. Petkovi\'{c}, S. D\v{z}eroski, D. Kocev \at
              Jo\v{z}ef Stefan Institute, Jamova 39, 1000 Ljubljana, Slovenia \\
              Jo\v{z}ef Stefan International Postgraduate School, Jamova 39, 1000 Ljubljana, Slovenia \\
              Tel.: +386-1477-3635\\
              \email{\{matej.petkovic,saso.dzeroski,dragi.kocev\}@ijs.si}           
}


\maketitle

\begin{abstract}
The data made available for analysis are becoming more and more complex along several directions: high dimensionality, number of examples and the amount of labels per example. This poses a variety of challenges for the existing machine learning methods: coping with dataset with a large number of examples that are described in a high-dimensional space and not all examples have labels provided. For example, when investigating the toxicity of chemical compounds there are a lot of compounds available, that can be described with information rich high-dimensional representations, but not all of the compounds have information on their toxicity. To address these challenges, we propose semi-supervised learning of feature ranking. The feature rankings are learned in the context of classification and regression as well as in the context of structured output prediction (multi-label classification, hierarchical multi-label classification and multi-target regression). To the best of our knowledge, this is the first work that treats the task of feature ranking within the semi-supervised structured output prediction context. More specifically, we propose two approaches that are based on tree ensembles and the Relief family of algorithms. The extensive evaluation across 38 benchmark datasets reveals the following: Random Forests perform the best for the classification-like tasks, while for the regression-like tasks Extra-PCTs perform the best, Random Forests are the most efficient method considering induction times across all tasks, and semi-supervised feature rankings outperform their supervised counterpart across a majority of the datasets from the different tasks.

\end{abstract}

\section{Introduction}
\label{intro}
\begin{sloppypar}
In the era of massive and complex data, predictive modeling is undergoing some significant changes. Since data are becoming ever more high dimensional, i.e., the target attribute potentially depends on a large number of descriptive attributes, there is a need to provide better understanding of the importance or {\it relevance} of the descriptive attributes for the target attribute. This is achieved through the task of feature ranking \cite{Guyon:JMLR:2003,Jong:PKDD:2004,Nilsson:JMLR:2007,petkomat:mtr}: the output of a feature ranking algorithm is a list (also called a {\it feature ranking}) of the descriptive attributes ordered by their relevance to the target attribute. The obtained feature ranking can then be used in two contexts: (1) to better understand the relevance of the descriptive variables for the target variable or (2) as a frequent pre-processing step to reduce the number of descriptive variables. By performing the latter, not only the computational complexity of building a predictive model later on is decreased, but at the same time, the models that use a lesser number of features are easier to explain and understand which is of high importance in a variety of application domains such as medicine \cite{medicine2,medicine1,medicine3}, life sciences \cite{Grissa16:jrnl,Saeys07:jrnl,Tsagris18:jrnl} and ecological modeling \cite{BHARDWAJ2018139,GALELLI201433,Zhou18:jrnl}.
\end{sloppypar}

Another aspect of massiveness is the number of examples in the data.
However, for some problems such as sentiment analysis of text, e.g., tweets \cite{petra},
or determining properties of new chemical compounds \cite{DiMasi}, e.g., in QSAR (quantitative structure activity relationship) studies (which is one of the considered datasets in the experiments), one can only label a limited quantity of data, since labeling demands a lot of human effort and time (labelling tweets), or is expensive (performing wet lab QSAR experiments). Since the cases where many examples remain unlabeled are not that rare, advances in predictive modeling have brought us to the point where we can make use of them. In this work, we focus on semi-supervised learning (SSL) techniques that handle data where some examples are labeled and some are not (as opposed to supervised learning (SL) where all examples are labeled).
Another direction of research goes into weakly supervised learning \cite{weak} where all examples may be labeled but (some) labels may be inaccurate or of a lower quality.

\begin{sloppypar}
The SSL approaches are all based on the assumption that the target values are well-reflected in the structure of the data, i.e.,
\begin{assumption}[Clustering Hypothesis]\label{lab:ch}
    Clusters of data examples (as computed in the descriptive space) well resemble the distribution of target values.
\end{assumption}
If the clustering hypothesis is satisfied, then a SSL algorithm that can make use of unlabeled data, may outperform the classical SL algorithms that simply ignore them.
This holds for predictive modeling tasks \cite{jurica:phd,ssl-intro}, and as we show in this work, for feature ranking tasks also.
\end{sloppypar}

In addition to the massiveness, the complexity of the data is also increasing. Predictive modeling is no longer limited to the standard classification and regression,
but also tackles their generalizations. For example, in classification, the target variable may take only one of the possible values, for each example in the data.
On the other hand, problems such as automatic tagging (e.g., the \texttt{Emotions} dataset (see Sec.~\ref{sec:data}) where the task is to determine emotions that a given musical piece carries)
allow for more than one label per example (e.g., a song can be \textit{sad} and \textit{dramatic} at the same time). A further generalization of this problem is hierarchical multi-label classification, where the possible labels are organized into a hierarchy, such as the one in Fig.~\ref{fig:hmlc-example}, which shows animal-related labels.
If a model labels an example as \emph{koala}, it should also label it with the generalizations of this label, i.e., \emph{Australian} and \emph{animal}.

Similarly, the task of regression can be generalized to multi-target regression, i.e., predicting more than one numeric variable at the same time, e.g., predicting canopy density and height of trees in forests (the \texttt{Forestry} dataset in Sec.~\ref{sec:data}).

The main motivation for the generalized predictive modeling tasks is that considering all the target variables at the same time may exploit the potential interactions among them
which are ignored when one predicts every variable separately. Moreover, building a single model for all targets can dramatically lower the computational costs.

In many cases, the data are at the same time semi-supervised (has missing), high dimensional and has a structured target, as for example in gene function prediction:
Labeling genes with their functions is expensive (semi-supervision), the genes can be described with a large number of variables (high dimensionality),
and the functions are organized into a hierarchy (structured target). Thus, designing feature ranking algorithms that i) can use unlabeled data, and ii) can handle
a variety of target types, including structured ones, is a relevant task that we address in this work. To the best of our knowledge, this is the first work that treats jointly the task of feature ranking in the context of semi-supervised learning for structured outputs.

\begin{sloppypar}
{\it We propose two general feature ranking approaches}. In the first approach, a ranking is computed from an ensemble of predictive clustering trees \cite{Kocev:Journal:2013,Blockeel98:phd},
adapted to structured outputs and SSL \cite{jurica:phd}, whereas the second approach is based on the distance-based Relief family of algorithms \cite{kira:relief}.
An initial study, investigated the performance of the ensemble-based approach in the classification task \cite{ssl-fr-stc}. In this work, we substantially extend our previous study in several directions:
\end{sloppypar}
\begin{enumerate}
    \item Additional datasets for classification are considered.
    \item Additional four tasks are considered (multi-label and hierarchical multi-label classification, single- and multi-target regression),
    and the ensemble-based feature ranking methods are evaluated in these cases.
    \item The Relief family of algorithms is extended to SSL, and evaluated for all five tasks
    (in comparison to the ensemble-based feature ranking methods).
\end{enumerate}

The rest of the paper is organized as follows.
In Sec.~\ref{sec:preliminaries}, we give the formal definitions of the different predictive modeling tasks, and introduce the notation.
Sec.~\ref{sec:related} surveys the related work, whereas Secs.~\ref{sec:ensemble-scores} and \ref{sec:relief-scores} define the ensemble-based and Relief-based feature importance scores,
respectively. Sec.~\ref{sec:setup} fully describes the experimental setup. We present and discuss the results in Sec.~\ref{sec:results}, and conclude with Sec.~\ref{sec:conclusions}.

The implementation of the methods, as well as the results are available at \url{http://source.ijs.si/mpetkovic/ssl-ranking}.

\section{Preliminaries}\label{sec:preliminaries}

{\bf Basic notation.} The data $\dataset{}$ consist of examples $(\bm{x}, \bm{y})$, where $\bm{x}$ is a vector of values of $D$ descriptive variables (features),
and $\bm{y}$ is the value of the target variable(s). The domain $\mathcal{X}_i$ of the feature $x_i$ is either numeric, i.e., $\mathcal{X}_i\subseteq \mathbb{R}$,
or categorical, i.e., it is a finite set of categorical values, e.g., $\mathcal{X}_i = \{A, B, AB, 0\}$ if a feature describes blood type.
Both numeric and categorical types are considered primitive \emph{unstructured} types. The domain $\mathcal{Y}$ of the target variable depends on the predictive modeling task at hand.
In this paper, we consider five tasks, two having unstructured, and three having structured target data types.

{\bf Regression (STR).} In this case, the target is a single numeric variable. Since we later consider also its generalization (multi-target regression), we refer to this task as single-target regression (STR).

{\bf Multi-target regression (MTR).} Here, the target variable is a vector with $T$ numeric variables as components, i.e., $\yDomain{}\subseteq\mathbb{R}^T$. Equivalently,
we can define MTR as having $T$ numeric targets, hence the name. In the special case of $T = 1$, MTR boils down to STR.

{\bf Classification.} In this case, the target is categorical. Since the algorithms considered in this paper can handle any classification task,
we do not distinguish between binary ($|\yDomain{}| = 2$) and multi-class classification ($|\yDomain{}| > 2$).

{\bf Multi-label classification (MLC).} The target domain  is a power set $\mathcal{P}(\labelset{})$ of some set $\labelset{}$ of categorical values, whose elements
are typically referred to as labels. Thus, the target values are sets. Typically, the target value $\bm{y}$ of the example $(\bm{x}, \bm{y})$ is referred to as a set of labels that are relevant for this example. The sets $\bm{y}$ can be of any cardinality, thus the labels are not mutually exclusive, as is the case with the task of (standard) classification.

\begin{figure}[!htb]
\centering
\includegraphics[width=.4\textwidth]{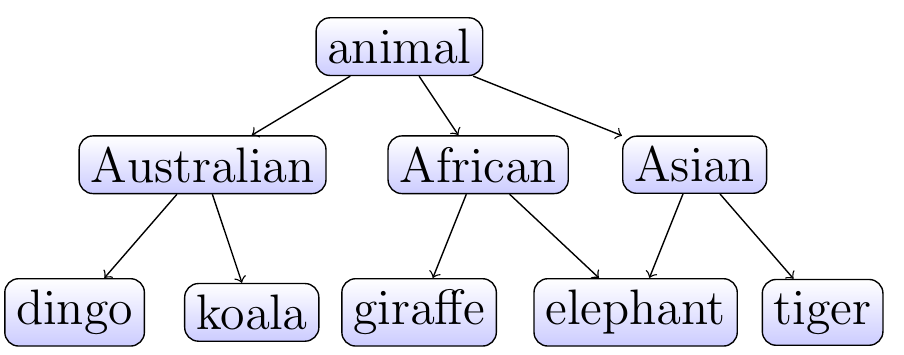}
\caption{An exemplary hierarchy of animal related labels.}
\label{fig:hmlc-example}
\end{figure}

{\bf Hierarchical multi-label classification (HMLC).} This is a generalization of MLC where the domain is again a power set of some label set $\labelset{}$,
which, additionally, is now partially-ordered via some ordering $\prec$. An exemplary hierarchy (of animal-related labels), which results from such an ordering is shown in the corresponding Haase diagram in Fig.~\ref{fig:hmlc-example}.

If $\ell_1\prec \ell_2$, the label $\ell_1$ is predecessor of the label $\ell_2$. If, additionally, there is no such label $\ell$,
such that $\ell_1 \prec \ell \prec \ell_2$, we say that $\ell_1$ is a parent of $\ell_2$. 
If a label does not have any parents, it is called a root.
A hierarchy can be either tree-shaped, i.e., every label has at most one parent, or it can be directed acyclic graph (DAG).
Since the label \texttt{elephant} has two parents (\texttt{African} and \texttt{Asian}), the hierarchy in Fig.~\ref{fig:hmlc-example} is not a tree.

Regarding predictive modeling, the ordering results in a hierarchical constraint, i.e., if a label $\ell$ is predicted to be relevant for a given example,
then, also its predecessors must be predicted relevant, e.g., if a given example is \texttt{koala}, it must also be \texttt{Australian} and \texttt{animal}.

In the cases of MLC and HMLC, each set of relevant labels $S\subseteq \labelset{}$ is conveniently represented by the 0/1 vector $\bm{s}$
of length $|\labelset{}|$, whose $j$-th component equals one if and only if $\ell_j\in S$. Thus, we will also use the notation $T = |\labelset{}|$.

{\bf Semi-supervised learning (SSL).} The unknown target values will be denoted by question marks ($?$). If the target value of the example is known,
we say that the example is labeled, otherwise the example is unlabeled. This applies to all types of targets and is not to be confused with the labels in the tasks of MLC and HMLC.

\section{Related Work}\label{sec:related}
\begin{sloppypar}
In general, feature ranking methods are divided into three groups \cite{vir:stanczykJainPregled}. \emph{Filter} methods do not need any underlying predictive model to compute the ranking. \emph{Embedded} methods compute the ranking directly from some predictive model. \emph{Wrapper} methods are more appropriate for feature selection, and build many predictive models which guide the selection.

Filters are typically the fastest but can be myopic, i.e., can neglect possible feature interactions, whereas the embedded methods are a bit slower, but can additionally serve as an explanation of the predictions of the underlying model. The prominence of the feature ranking reflects in numerous methods solving this task in the context of classification and STR \cite{Guyon:JMLR:2003,vir:stanczykJainPregled}, however, the territory of feature ranking for SSL is mainly uncharted, especially when it comes to structured output prediction.

An overview of SSL feature ranking methods for classification and STR is given in \cite{Sheikhpour:ssl-overview}. However, the vast majority of the methods described there are either supervised or unsupervised (ignoring the labels completely). An exception is the SSL Laplacian score \cite{doquire:ssl-reg}, applicable to the STR problems.
\end{sloppypar}

This method is a filter and stems from graph theory. It first converts a dataset into a graph,
encoded as a weighted incidence matrix whose weights correspond to the distances among the examples in the data.
The distances are measured in the descriptive space but more weight is put on the labeled examples.
One of the drawbacks of the original method is that it can only handle numeric features. Our modification that overcomes this is described in Sec.~\ref{sec:considered-methods}.

For structured output prediction in SSL, we could not find any competing feature ranking methods. Our ensemble-based scores belong to the group of embedded methods, and crucially depend on ensembles of SSL predictive clustering trees (PCTs) \cite{jurica:phd}. We thus describe bellow SSL PCTs and ensembles thereof.

\subsection{Predictive clustering trees}\label{sec:pcts}
PCTs are a generalization of standard decision trees. They can handle various structured output prediction tasks and have been recently adapted to SSL \cite{jurica:phd}.
This work considers the SSL of PCTs for classification, STR, MTR \cite{jurica-mtr}, MLC, and HMLC.

For each of these, one has to specify the impurity function $\impurity{}$ that is used in the best test search (Alg.~\ref{alg:bestTest}),
and the prototype function $\prototype{}$ that creates the predictions in the leaf nodes.
After these two are specified, a PCT is induced in the standard top-down-tree-induction manner.

Starting with the whole dataset $\datatrain{}$, we find the test (Alg.~\ref{alg:pct}, line~\ref{alg:pct:search}) that greedily splits the data so that the heuristic score of the test, i.e., the decrease of the impurity $\impurity{}$ of the data after applying the test, is maximized. For a given test, the corresponding decrease is computed in line~\ref{alg:heurCom} of Alg.~\ref{alg:bestTest}.

If no useful test is found, the algorithm creates a leaf node and computes a leaf node with the prediction (Alg.~\ref{alg:pct}, line~\ref{alg:pct:leaf}). Otherwise, an internal node $\node{}$ with the chosen test is constructed, and the PCT-induction algorithm is recursively called on the subsets in the partition of the data, defined by the test.
The resulting trees become child nodes of the node $\node{}$ (Alg~\ref{alg:pct}, line~\ref{alg:pct:internal}).
\vskip-1.5em
\noindent\begin{minipage}[!T]{0.43\textwidth}
    \begin{algorithm}[H]
    \caption{PCT($E$)}\label{alg:pct}
     \begin{algorithmic}[1]
         \STATE $(t^*,h^*,\mathcal{P}^*) = \mathrm{BestTest}(E)$\label{alg:pct:search}
         \IF{$t^* = \mathit{none}$}
            \STATE \textbf{return} $\mathit{Leaf}(\prototype{}(E))$\label{alg:pct:leaf}
        \ELSE
             \FOR{\textbf{each} $E_i \in \mathcal{P^*}$}
                \STATE $\mathit{tree}_i$ = PCT($E_i$)
             \ENDFOR
             \STATE \textbf{return} $\mathit{Node}(t^*,\;\bigcup_i \{\mathit{tree}_i\})$\label{alg:pct:internal}
         \ENDIF
    \end{algorithmic}
    \end{algorithm}
\end{minipage}
\begin{minipage}[!T]{0.55\textwidth}
\begin{algorithm}[H] 
    \caption{$\mathrm{BestTest}(E)$}\label{alg:bestTest}
 \begin{algorithmic}[1]
     \STATE $(t^*,h^*,\mathcal{P}^*) = (\mathit{none},0,\emptyset)$
     \FOR{\textbf{each} test $t$} \label{alg:btest1}
         \STATE $\mathcal{P} = $ partition induced by $t$ on $E$\label{alg:partition}
         \STATE $h = |E|\impurity{}(E) -\sum_{E_i \in \mathcal{P}} |E_i| \impurity{}(E_i)$\label{alg:heurCom}
        \IF{$h > h^*$} \label{alg:icv}
            \STATE $(t^*,h^*,\mathcal{P}^*) = (t,h,\mathcal{P})$ \label{alg:btest2}
        \ENDIF
     \ENDFOR
     \STATE \textbf{return} $(t^*,h^*,\mathcal{P}^*)$
 \end{algorithmic}
 \end{algorithm}
\end{minipage}
\vskip1em
\noindent The impurity functions for a given subset $E\subseteq \datatrain{}$ in the considered tasks are defined as weighted averages of the feature impurities $\impurity{}(E, x_i)$, and target impurities $\impurity{}(E, y_j)$.

For nominal variables $z$, the impurity is defined in terms of the Gini Index $\gini{}(E, z) = 1 - \sum_v p_E^2(v)$, where the sum goes over the possible values $v$ of the variable $z$, and $p_E(v)$ is the relative frequency of the value $v$ in the subset $E$.
In order not to favoritize any variable a priori, the impurity is defined as the normalized Gini value, i.e., 
$\impurity{}(E, z) = \gini{}(E, z) / \gini{}(\datatrain{}, z)$. This applies to nominal features and the target in classification.

For numeric variables $z$, the impurity is defined in terms of their variance $\var{}(E, z)$, i.e., $\impurity{}(E, z) = \var{}(E, z) / \var{}(\datatrain{}, z)$.
This applies to numeric features and targets in other predictive modeling tasks, since the sets in MLC and HMLC are also represented by 0/1 vectors.
However, note that computing the Gini-index of a binary variable is equivalent to computing the variance of this variable if the two values are mapped to $0$ and $1$.
When computing the single-variable impurities, missing values are ignored.

In a fully-supervised scenario, the impurity of data is measured only on the target side. However, the majority of target values
may be missing in the semi-supervised case. Therefore, for SSL, also the features are taken into account when calculating the impurity, which is defined as
\begin{equation}
    \label{eq:impurity}
    \impurity{}(E) = w\cdot \frac{1}{T} \sum_{j = 1}^T \alpha_j \impurity{}(E, y_j) + (1 - w)\cdot \frac{1}{D} \sum_{i = 1}^D \beta_i \impurity{}(E, x_i)\text{,}
\end{equation}
where the level of supervision is controlled by the user-defined parameter $w\in [0, 1]$. Setting it to $1$ means fully-supervised
tree-induction (and consequently ignoring unlabeled data). The other extreme, i.e., $w = 0$, corresponds to fully-unsupervised
tree-induction (also known as clustering). The dimensional weights $\alpha_j$ and $\beta_i$ are typically all set to $1$, except for HMLC where
$\alpha_i = 1$ for the roots of the hierarchy, and $\alpha_i = \alpha \cdot \mathit{mean}(\text{parent weights})$ otherwise, where $\alpha\in (0, 1)$ is a user-defined parameter. A MLC problem is considered a HMLC problem where all labels are roots.

The prototype function returns the majority class in the classification case, and the per-component mean $[\bar{y}_1, \dots, \bar{y}_T]$ of target vectors otherwise.
In all cases, the prototypes (predictions) are computed from the training examples in a given leaf. In the cases of MLC and HMLC, the values $\bar{y}_j$ can be additionally thresholded to obtain the actual subsets, i.e., $\hat{\bm{y}} = \{\ell_j \mid \bar{y}_j \geq \vartheta, 1\leq j\leq T\}$, where
taking $\vartheta = 0.5$ corresponds to computing majority values of each label.

\subsection{Ensemble methods}
To obtain a better predictive model, more than one tree can be grown, for a given dataset, which results in an ensemble of trees.
Predictions of an ensemble are averaged predictions of trees (or, in general, arbitrary base models) in the ensemble.
However, a necessary condition for an ensemble to outperform its base models is, that the base models are diverse \cite{Hansen90:jrnl}.
To this end, some randomization must be introduced into the tree-induction process, and three ways to do so have been used \cite{jurica:phd}.

{\bf Bagging.} When using this ensemble method, instead of growing the trees using $\datatrain{}$, a bootstrap replicate of $\datatrain{}$ is independently created for each tree, and used for tree induction.

{\bf Random Forests (RFs).} In addition to the mechanism of Bagging, for each internal node of a given tree, only a random subset (of size $D' < D$) of all features is considered when searching for the best test, e.g., $D' = \mathit{ceil}(\sqrt{D})$.

{\bf Extremely Randomized PCTs (ETs).} As in Random Forests, a subset of features can be considered in every internal node (this is not a necessity), but additionally, only one test per feature is randomly chosen and evaluated. In contrast to Random Forests (and Bagging), the authors of original ETs did not use bootstrapping \cite{geurts:extraT}. However, previous experiments \cite{ssl-fr-stc} showed that it is beneficial to do so when the features are (mostly) binary, since otherwise ets can offer only one possible split and choosing one at random has no effect.

\section{Ensemble-Based Feature Ranking}\label{sec:ensemble-scores}
The three proposed importance scores can be all computed from a single PCT, but to stabilize the scores, they are rather computed from an ensemble:
Since the trees are grown independently, the variance of each score $\importance{}(x_i)$ decreases linearly with the number of trees.

Once an ensemble (for a given predictive modeling task) is built,
we come to the main focus of this work: Computing a feature ranking out of it. There are three ways to do so: Symbolic \cite{petkomat:mtr},
Genie3 \cite{petkomat:mtr} (its basic version (for standard classification and regression) was proposed in \cite{genie3}),
and Random Forest score \cite{petkomat:mtr} (its basic version was proposed in \cite{Breiman01a:jrnl}):
\begin{eqnarray}
\label{eq:symbolic}
\importance_\text{SYMB}(x_i) &=& \frac{1}{|\forest|} \sum_{\tree\in\forest} \sum_{\node\in\tree(x_i)} |E(\node{})| / |\datatrain{}|\text{,} \\
\label{eq:genie}
\importance_\text{GENIE3}(x_i) &=& \frac{1}{|\forest|} \sum_{\tree \in \forest} \sum_{\node \in \tree(x_i)} h^*(\node)\text{,}\\
\label{eq:rf}
\importance_\text{RF}(x_i) &=& \frac{1}{|\forest|}\sum_{\tree \in \forest}
\frac{\error(\text{OOB}_\tree^i) - \error(\text{OOB}_\tree)}{\error(\text{OOB}_\tree)}\text{.}
\end{eqnarray}
Here, $\forest{}$ is an ensemble of trees $\tree{}$, $\tree{}(x_i)$ is the set of the internal nodes $\node{}$ of a tree $\tree{}$ where the feature $x_i$ appears in the test, $E(\node{}) \subseteq\datatrain{}$ is the set of examples that reach the node $\node{}$,
$h^*$ is the heuristic value of the chosen test, $\error{}(\text{OOB}_\tree{})$ is the value of the error measure $\error{}$,
when using $\tree{}$ as a predictive model for the set $\text{OOB}_\tree{}$ of the out-of-bag examples for a tree $\tree{}$, i.e., examples that were not chosen into the bootstrap replicate, thus not seen during the induction of $\tree{}$. Similarly, $\error{}(\text{OOB}_\tree{}^i)$ is the value of the error measure $\error{}$ on the $\text{OOB}_\tree{}$ with randomly permuted values of the feature $x_i$.

Thus, Symbolic and Genie3 ranking take into account node statistics: The Symbolic score's award is proportional to the number of examples that reach this node, while Genie3 is more sophisticated and takes into account also the heuristic value of the test (which is proportional to $|E(\node{})|$, see Alg.~\ref{alg:bestTest}, line \ref{alg:heurCom}.

The Random Forest score, on the other hand, measures to what extent noising, i.e., permuting, the feature values decreases the predictive performance of the tree. In Eq.~\eqref{eq:rf}, it is assumed that $\error{}$ is a loss, i.e., lower is better as is the case, for example, in the regression problems where (relative root) mean squared errors are used. Otherwise, e.g., for classification tasks and the $F_1$ measure, the importance of a feature is defined as $-\importance{}_\text{RF}$ from Eq.~\eqref{eq:rf}.
Originally, it was designed to explain the predictions of the RFs ensemble \cite{Breiman01a:jrnl} (hence the name), but it can be used with any predictive model.
However, trees are especially appropriate, because the predictions can be obtained fast, provided the trees are balanced.

\subsection{Ensemble-based ranking for SSL structured output prediction}
The PCT ensemble-based feature ranking methods for different structured output prediction (SOP) tasks have been introduced by \cite{petkomat:mtr,acta-hun-hmlc}, and evaluated for different SL SOP tasks. In this case, PCTs use a heuristic based on the impurity reduction on the target space,
as defined by Eq.~\eqref{eq:impurity}, in a special case when $w = 1$. As for SSL, the general case of Eq.~\eqref{eq:impurity} applies.
Once we have SSL PCTs, the ensemble-based feature ranking methods technically work by default. They have been evaluated in the case of SSL classification. However, they have not been evaluated on STR and SOP tasks.

\subsection{Does the ensemble method matter?}
From Eqs.~\eqref{eq:symbolic}--\eqref{eq:rf}, it is evident that all three feature ranking scores can in theory be computed from a single tree, and averaging them
over the trees in the ensemble only gives a more stable estimate of $\mathbb{E}[\importance{}(x_i)]$.
However, one might expect that bagging, RFs and ETs on average yield the same order of features (or even the same importance values)
since the latter two are more randomized versions of the bagging method. Here, we sketch a proof that this is not (necessarily) the case.

One of the cases when the expected orders of features are equal, is a dataset where each of the two binary features
$x_1$ and $x_2$ completely determine the target $y$, e.g., $y= x_1$ and $y = 1 - x_2$, and the third feature is effectively random noise.
It is clear that the expected values of the importances are in all cases $\importance{}(x_1) = \importance{}(x_2) > \importance{}(x_3)$.

One of the cases where bagging gives rankings different from those of RFs, is a dataset where knowing the values of ranking pairs $(x_1, x_2)$
and $(x_3, x_i)$, for $4\leq i\leq D$ again completely reconstructs the target value $y$, and $h(x_1) > h(x_i) > \max\{h(x_2), h(x_3)\}$, for $i\geq 4$.
In this case, bagging will first choose $x_1$ and then $x_2$ in the remaining two internal nodes of the tree, so $x_1$ and $x_2$ would be the most important features.
On the other hand, RFs with $D' = 1$ and $D$ sufficiently large, will in the majority of the cases first choose one of the features $x_i$, $i\geq 4$,
and then, sooner or later, $x_3$. Unlike in the bagging-based ranking, $x_3$ is now more important than $x_1$.

\subsection{Time complexity}\label{sec:times-ensemble}
In predictive clustering, the attributes in the data belong to three (not mutually exclusive) categories: i) Descriptive attributes are those that can appear in tests of internal nodes of a tree, ii) Target attributes are those for which predictions in leaf nodes of a tree are made, and iii) Clustering attributes are those that are used in computing the heuristic when evaluating the candidate tests. Let their numbers be $D$, $T$ and $C$, respectively, and let $M$ be the number of examples in $\datatrain{}$.
Note that in the SSL scenario (if $w\notin \{0, 1\}$), we  have the relation $C = D + T$.
Assuming that the trees are balanced, we can deduce that growing a single semi-supervised tree takes $\mathcal{O}(M D' \log M (\log M + C))$ \cite{jurica:phd}.

After growing a tree, ranking scores are updated in $\mathcal{O}(M)$ time (where $M$ is the number of internal nodes) for the Symbolic and Genie3 score, whereas updating the Random Forest scores takes $\mathcal{O}(D M \log M)$. Thus, computing the feature ranking scores does not change the $\mathcal{O}$-complexity of growing a tree,
and we can compute all the rankings from a single ensemble.
Thus, growing an ensemble $\forest{}$ and computing the rankings takes $\mathcal{O}(|\forest{}| M D' \log M (\log M + C))$.

\section{Relief-based Feature Ranking}\label{sec:relief-scores}
\begin{sloppypar}
The \textsc{Relief} family of feature ranking algorithms does not use any predictive model.
Its members can handle various predictive modeling tasks, including classification \cite{kira:relief}, regression \cite{kononenko:relief},
MTR \cite{petkomat:mtr}, MLC \cite{petkomat:mlc:relief,reyes}, and HMLC \cite{acta-hun-hmlc}. The main intuition behind Relief is the following:
the feature $x_i$ is relevant if the differences in the target space between two neighboring examples are notable if and only if the differences in the feature values of $x_i$ between these two examples are notable.

\subsection{Supervised Relief}
More precisely, if $\bm{r} = (\bm{x}^1, \bm{y}^1)\in \datatrain{}$ is randomly chosen,
and $\bm{n} = (\bm{x}^2, \bm{y}^2)$ is one of its nearest $k$ neighbors, then the computed importances  $\importance{}_\text{Relief}(x_i)$ of the Relief algorithms equal the estimated value of
\end{sloppypar}
\begin{equation}\label{eqn:relief}
      P_1 - P_2 = P(\bm{x}_i^1 \neq \bm{x}_i^2 \mid  \bm{y}^1 \neq \bm{y}^2) - P(\bm{x}_i^1 \neq \bm{x}_i^2 \mid  \bm{y}^1 = \bm{y}^2)\text{,}
\end{equation}
where the probabilities are modeled by the distances between $\bm{r}$ and $\bm{n}$ in appropriate subspaces. For the descriptive space $\mathcal{X}$ spanned by the domains $\mathcal{X}_i$ of the features $x_i$, we have
\begin{equation}
\label{eqn:metric}
d_\mathcal{X}(\bm{x}^1, \bm{x}^2) = \frac{1}{F}\sum_{i = 1}^F d_i(\bm{x}^1, \bm{x}^2);\;\;
d_i(\bm{x}^1, \bm{x}^2)=
\begin{cases}
\;\bm{1}[\bm{x}_i^1 \neq \bm{x}_i^2]&: \mathcal{X}_i \nsubseteq \mathbb{R}\\
\frac{|\bm{x}_i^1 -\bm{x}_i^2|}{\max\limits_{\bm{x}} \bm{x}_i - \min\limits_{\bm{x}}\bm{x}_i} &: \mathcal{X}_i\subseteq \mathbb{R}
\end{cases}
\end{equation}
where $\bm{1}$ denotes the indicator function. The definition of the target space distance $d_\mathcal{Y}$ depends on the target domain. In the cases of classification and MTR, the categorical and numeric part of the definition $d_i$ in Eq.~\eqref{eqn:metric} apply, respectively. Similarly, in multi-target regression, $d_\mathcal{Y}$ is the analogue of $d_\mathcal{X}$ above.

In the cases MLC and HMLC, we have more than one option for the target distance definition \cite{petkomat:mlc:relief}, but in order to be as consistent as possible
with the STR and MTR cases, we use the Hamming distance between the two sets. Recalling that sets $S\subseteq\labelset{}$ are presented as 0/1 vector $\bm{s}$ (Sec.~\ref{sec:preliminaries}), the Hamming distance $d_\mathcal{Y}$ is defined as
\begin{equation}
    \label{eqn:hamming}
    d_\mathcal{Y}(S^1, S^2) = \gamma \sum_{i = 1}^{|\labelset{}|} \alpha_i \bm{1}[\bm{s}_i^1 \neq \bm{s}_i^2]
\end{equation}
where the weights $\alpha_i$ are based on the hierarchy and are defined as in Eq.~\eqref{eq:impurity}, and $\gamma$ is the normalization factor that assures
that $d_{\yDomain{}}$ maps to $[0, 1]$. It equals $\frac{1}{|\labelset{}|}$ in the MLC case, and depends on the data in the HMLC case \cite{acta-hun-hmlc}.

To estimate the conditional probabilities $P_{1, 2}$ from Eq.~\eqref{eqn:relief}, they are first expressed in the unconditional form, e.g., $P_1 = P(\bm{x}_i^1 \neq \bm{x}_i^2 \land  \bm{y}^1 \neq \bm{y}^2) / P(\bm{y}^1 \neq \bm{y}^2)$. Then, the numerator is modeled as the product $d_i d_\mathcal{Y}$, whereas the nominator
is modeled as $d_\mathcal{Y}$. The probability $P_2$ is estimated analogously.

\subsection{Semi-supervised Relief}
In the SSL version of the above tasks, we have to resort to the predictive clustering paradigm, using descriptive and clustering attributes instead of descriptive and target ones. More precisely, the descriptive distance is defined as above. As for the clustering distance, it equals $d_\mathcal{Y}$ when the target value of both $\bm{y}^1$ and $\bm{y}^2$ are known, and equals $d_\mathcal{X}$ otherwise. The contribution of each pair to the estimate of probabilities is weighted according to their distance to the labeled data. The exact description of the algorithm is given in Alg.~\ref{alg:relief}.
\begin{algorithm}
	\caption{SSL-Relief($\datatrain{}$, $m$, $k$, $[w_0, w_1]$)}
	\label{alg:relief}  
	\begin{algorithmic}[1]
		\STATE{$\textbf{imp} = $ zero list of length $D$}
		\STATE{$\bm{P}_\text{diffAttr, diffCluster}$, $\bm{P}_\text{diffAttr} = $ zero lists of length $D$}
		\STATE{$P_\text{diffCluster}= 0.0$}
		\STATE{$\bm{w} = \mathit{computeInstanceInfluence(\datatrain{}, w_0, w_1)}$}\label{alg:relief:influence}\label{line:influence}
		\STATE{$s = 0$\hfill \# sum of weights of the pairs, used in normalization}
		\FOR{iteration $= 1, 2,\dots, m$}
			\STATE{$\bm{r} = $ random example from $\dataset$}\label{line:rndEx}
			\STATE{$\bm{n}_1, \bm{n}_2, \dots, \bm{n}_k =$ $k$ nearest neighbors of $\bm{r}$}\label{line:knn}
			\FOR{$\ell = 1,2, \dots, k $}
			    \STATE{$w = \bm{w}[\bm{r}] \cdot \bm{w}[\bm{n}_\ell]$}
			    \STATE{$s \pluseq w$}
			    \IF{$\bm{r}$ and $\bm{n}_\ell$ are labeled}
			        \STATE{$d_\text{cluster} = d_\mathcal{Y}\big(\bm{r}, \bm{n}_\ell\big)$}
			    \ELSE
			        \STATE{$d_\text{cluster} = d_\mathcal{X}\big(\bm{r}, \bm{n}_\ell\big)$}
			    \ENDIF
				\STATE{$P_\text{diffCluster} \pluseq w\; d_\text{cluster}\big(\bm{r}, \bm{n}_\ell\big)$}
				\FOR{$i = 1, 2, \dots, D$}\label{line:updateStart}
					\STATE{$\bm{P}_\text{diffAttr}[i] \pluseq w\; d_i\big(\bm{r}, \bm{n}_\ell\big)$}
					\STATE{$\bm{P}_\text{diffAttr, diffCluster}[i] \pluseq w\; d_i\big(\bm{r}, \bm{n}_\ell\big)\, d_\text{cluster}\big(\bm{r}, \bm{n}_\ell\big) $}\label{line:updateEnd}
				\ENDFOR
			\ENDFOR
		\ENDFOR
		\FOR{$i = 1,2, \dots, D$}
			\STATE{$\textbf{imp}[i] = \frac{\bm{P}_\text{diffAttr, diffCluster}[i]}{P_\text{diffCluster}}
			- \frac{\bm{P}_\text{diffAttr}[i] - \bm{P}_\text{diffAttr, diffCluster}[i]}{s - P_\text{diffCluster}}$}\label{line:weights}
		\ENDFOR
		\RETURN{$\textbf{imp}$}
	\end{algorithmic}  
\end{algorithm}

SSL-Relief takes as input the standard parameters ($\datatrain{}$, the number of iterations $m$, and the number of Relief neighbors $k$),
as well as the interval $[w_0, w_1]\subseteq [0, 1]$, which the influence levels of $\bm{r}$-$\bm{n}$ pairs are computed from (line \ref{alg:relief:influence}):
First, for every $(\bm{x}, \bm{y})\in\datatrain{}$, we find the distance $d_{\bm{x}}$ to its nearest labeled neighbor.
If $d = 0$, i.e., the value $\bm{y}$ is known, the influence $w$ of this example is set to $1$. Otherwise, the influence of the example is defined by a linear function $d\mapsto w(d)$ that goes through the points  $(\max_{(\bm{x}, ?)} d_{\bm{x}}, w_0)$ and $(\min_{(\bm{x}, ?)} d_{\bm{x}}, w_1)$. Thus, the standard regression version of Relief is obtained when no target values are missing.

\subsection{Time complexity}\label{sec:times-relief}
For technical reasons, the actual implementation of SSL-Relief does not follow the Alg.~\ref{alg:relief} word for word, and first computes all nearest neighbors.
This takes $\mathcal{O}(m M D)$ steps, since the majority of the steps in this stage is needed for computing the distances in the descriptive space.
We use the brute-force method, because it is, for the data at hand, still more efficient than, for example, k-D trees. Since the number of iterations is typically set to be a proportion of $M$ (in our case $m = M$), the number of steps is $\mathcal{O}(M^2 D)$. When computing the instances' influence (line \ref{line:influence}),
only the nearest neighbor of every instance is needed, so this can be done after the $K$-nearest neighbors are computed, within a negligible number of steps.

In the second stage, the probability estimates are computed and the worst-case time complexity is achieved when all examples are labeled since this is the case
when we have to additionally compute $d_{\yDomain{}}$ (otherwise, we use the stored distances $d_{\mathcal{X}}$).
The number of steps needed for a single computation od $d_{\yDomain{}}$ depends on the domain:
$\mathcal{O}(1)$ suffices for classification and STR, whereas $\mathcal{O}(T)$ steps are required in the MTR, MLC and HMLC cases.

The estimate updates themselves take $\mathcal{O}(D)$ steps per neighbor, thus, the worst case time complexity is
$\mathcal{O}(M^2 D + k M (T + D)) = \mathcal{O}(M^2 D + k M C)$ where $C = D + T$ is (again) the number of clustering attributes.

\section{Experimental Setup}\label{sec:setup}

In this section, we undertake to experimentally evaluate the proposed feature ranking methods. We do so by answering a set of experimental questions listed below. We then describe in detail how the experimental evaluation is carried out. 

\subsection{Experimental questions}
The evaluation is based on the following experimental questions:
\begin{enumerate}
    \item For a given ensemble-based feature ranking score, which ensemble method is the most appropriate?
    \item Are there any qualitative differences between the semi-supervised and supervised feature rankings?
    \item Can the use of unlabeled data improve feature ranking?
    \item Which feature ranking algorithm performs best?
\end{enumerate}

\subsection{Datasets}\label{sec:data}

All datasets are well-known benchmark problems that come from different domains. For classification, we have included five new datasets (those below the splitting line of  Tab.~\ref{tab:data:stc}), in addition to the previous ones \cite{ssl-fr-stc}.

Since MLC can be seen as a special case of HMLC with a trivial hierarchy, we show the basic characteristics of the considered MLC and HMLC problems in a single table (Tab.~\ref{tab:data:mlc}), separating the MLC and HMLC datasets by a line.

Similarly, the regression problems (for STR and MTR) are shown in Tab.~\ref{tab:data:r}.

The given characteristics of the data differ from tasks to task, but the last column of every table (CH) always gives the estimate of 
how well the clustering hypothesis (Asm.~\ref{lab:ch}) holds. For all predictive modeling tasks, this estimate is based on
$k$-means clustering \cite{k-means} or, more precisely, on the agreement between the distribution of the target values in these clusters.
The number of clusters was set to the number of classes in the case of classification, and to $8$ otherwise, i.e.,
the default Scikit Learn's \cite{scikit-learn} parameters are kept. The highest agreement of the five runs of the method is reported.

{\bf CH computation.} In the case of classification, the measure at hand is the Adjusted Random Index \cite{ari} (ARI) that we have already used earlier \cite{ssl-fr-stc}. It
computes the agreement between the classes that examples are assigned via clustering, and the actual class values.
The optimal value of ARI is $1$, whereas the value $0$ corresponds to the case when clustering
is independent of class distribution.

In the other cases, we compute the variance of each target variable, i.e., an actual target in the  STR and MTR case,
and a component of the 0/1 vector which a label set in the case of MLC and HMLC is represented by.
Let $\mathcal{C}$ be the set of the obtained clusters, i.e., $c\subseteq \dataset{}$, for each cluster $c\in\mathcal{C}$.
Then, for every target variable $y_j$, we compute $v_j = \sum_{c} p(c) \var{}(c, y_j) / \var{}(\dataset{}, y)$, i.e., the relative decrease of the variance after the clustering is applied, where $p(c)= |c|/|\dataset{}|$.
It can be proved (using the standard formula for the estimation of sample variance and some algebraic manipulation) that $v_j\leq 1$. Trivially, $v_j\geq 0$. We average the contributions $v_j$ over the target variables
to obtain the score $v$. In the case of HMLC, we use weighted average where the weights are proportional to the hierarchical weights $\alpha_i$, defined in Sec.~\ref{sec:pcts}. Finally, the tables report the values of $\mathit{CH} = 1 - v\in [0, 1]$, to make the value $1$ optimal.

\begin{table}[!b]
  \centering
  \caption{Basic properties of the classification datasets:
  number of examples $|\dataset{}|$, number of features $D$, number of classes (the $y$-domain size $|\yDomain{}|$), the proportion of examples in the majority class (MC), and the CH value.}
    \begin{tabular}{lrrrrr}
    \hline
    dataset & \multicolumn{1}{c}{$|\dataset{}|$} & \multicolumn{1}{c}{$D$} & \multicolumn{1}{c}{$|\yDomain{}|$} & \multicolumn{1}{l}{MC} & CH \\
    \hline
    Arrhythmia \cite{uci}           & 452  & 279& 16& 0.54& 0.02\\
    Bank  \cite{uci,moro}           & 4521 & 16 & 2 & 0.88& -0.00\\
    Chess  \cite{uci}               & 3196 & 36 & 2 & 0.52& 0.22\\
    Dis  \cite{openml}              & 3772 & 28 & 2 & 0.98& 0.00\\
    Gasdrift \cite{uci}             & 13910& 128& 6 & 0.22& 0.02\\
    Pageblocks  \cite{uci}          & 5473 & 10 & 5 & 0.90& 0.03\\
    Phishing  \cite{uci}            & 11055& 30 & 2 & 0.56& -0.00\\
    Tic-tac-toe \cite{uci}          & 958  & 9  & 2 & 0.65& 0.70\\
    \hline                                            
    Aapc \cite{aapc}                & 335  & 84 & 3 & 0.47& 0.34\\
    Coil2000  \cite{van2004bias}    & 9822 & 85 & 2 & 0.94& -0.00\\
    Digits  \cite{xu1992methods}    & 1797 & 64 & 10& 0.10& -0.00\\
    Pgp   \cite{levatic2013accurate}& 932  & 183& 2 & 0.52& 0.00\\
    Thyroid  \cite{uci}             & 3772 & 27 & 2 & 0.94& 0.01\\
    \hline
    \end{tabular}%
  \label{tab:data:stc}%
\end{table}%

\begin{table}[!b]
  \centering
  \caption{Basic properties of the MLC (above the line) and HMLC (below the line) datasets:
  number of examples $|\dataset{}|$, number of features $D$, number of labels $|\labelset{}|$, label cardinality (average number of labels per example) $\ell_c$,
  the depth of hierarchy, and the CH value.}
    \begin{tabular}{lrrrrrrr}
    \hline
    dataset & \multicolumn{1}{c}{$|\dataset{}|$} & \multicolumn{1}{c}{\;\;$D$} & \multicolumn{1}{c}{$|\labelset{}|$} &\multicolumn{1}{l}{\;\;$\ell_c$} & depth & shape   & CH \\
    \hline
    Bibtex \cite{KTV08}                         & 7395 & 1836 & 159& 2.4 & 1 & tree & 0.02   \\ 
    Birds \cite{birds}                          & 645  & 260  & 19 & 1.0 & 1 & tree & 0.05   \\ 
    Emotions \cite{emotions}                    & 593  & 72   & 6  & 1.9 & 1 & tree & 0.04   \\ 
    Genbase \cite{genbase}                      & 662  & 1185 & 27 & 1.3 & 1 & tree & 0.26   \\ 
    Medical \cite{medical}                      & 978  & 1449 & 45 & 1.3 & 1 & tree & 0.04   \\ 
    Scene \cite{scene}                          & 2407 & 294  & 6  & 1.1 & 1 & tree & 0.21   \\ 
    \hline
    Clef07a-is \cite{clef07a-is}       &11006& 80  &  96 & 3.0 & 3.0& tree& 0.05 \\  
    Ecogen \cite{ecogen}               & 1893& 138 &  56 & 15.5& 3.0& tree& 0.03 \\  
    Enron-corr \cite{enron-corr}       &1648 & 1001&  67 &  5.3& 3.0& tree& 0.03 \\  
    Expr-yeast-FUN \cite{Clare03:phd}  &3788 & 552 &  594&  8.9& 4.0& tree& 0.00 \\  
    Gasch1-yeast-FUN \cite{Clare03:phd}&3773 & 173 &  594&  8.9& 4.0& tree& 0.01 \\  
    Pheno-yeast-FUN \cite{Clare03:phd} &1592 & 69  &  594&  9.1& 4.0& tree& 0.00 \\  
    \hline
    \end{tabular}%
  \label{tab:data:mlc}%
\end{table}%

\begin{table}[!b]
  \centering
  \caption{Basic properties of the STR and MTR datasets:
  number of examples $|\dataset{}|$, number of features $D$, number of targets $T$, and the CH value.}
    \begin{tabular}{lrrrrr}
    \hline
    dataset & \multicolumn{1}{l}{examples} & \multicolumn{1}{c}{$D$} & \multicolumn{1}{l}{\;\;$T$} & \multicolumn{1}{l}{\;\;CH} \\
    \hline
    CHEMBL2850 \cite{openml}           & 1211 & 1024& 1 & 0.09 \\ 
    CHEMBL2973 \cite{openml}           & 1521 & 1024& 1 & 0.18 \\ 
    Mortgage  \cite{bilken}            & 1049 & 15  & 1 & 0.57 \\ 
    Pol  \cite{bilken}                & 5000 & 26  & 1 & 0.12 \\ 
    QSAR \cite{openml}                 & 2145 & 1024& 1 & 0.20 \\ 
    Treasury  \cite{bilken}            & 1049 & 15  & 1 & 0.54 \\ 
    \hline
    Atp1d \cite{Spyromitros}                & 337  & 411 & 6 & 0.49 \\ 
    CollembolaV2 \cite{Kampichler00:jrnl}         & 393  & 47  & 3 & 0.02 \\ 
    Edm1 \cite{Karalic97:jrnl}                & 154  & 16  & 2 & 0.23 \\ 
    Forestry-LIDAR-IRS \cite{Stojanova:msc}   & 2730 & 28  & 2 & 0.19 \\ 
    Oes10  \cite{Spyromitros}            & 403  & 298 & 16& 0.63 \\ 
    Scm20d \cite{Spyromitros}               & 8966 & 61  & 16& 0.16 \\ 
    Soil-quality  \cite{soil-q}        & 1944 & 142 & 3 & 0.07 \\ 
    \hline
    \end{tabular}%
  \label{tab:data:r}%
\end{table}%

\subsection{Parameter instantiation}
We parametrize the used methods as follows. The number of trees in the ensembles was set to $100$ \cite{Kocev:Journal:2013}.
The number of features that are considered in each internal node was set to $\sqrt{D}$ for RFs and $D$ for ETs \cite{geurts:extraT}. The optimal value of the level of supervision parameter $w$ for computing the ensembles of PCTs was selected by internal 4-fold cross-validation.
The considered values were $w\in\{0, 0.1, 0.2, \dots, 0.9, 1\}$.

The amount of supervision in SSL-Relief is adaptive, which allows for coarser set of values,
and we consider $w_{1, 2}\in\{0, 0.25, 0.5, 0.75, 1\}$ (where $w_1\leq w_2$). The considered numbers $k$ of Relief neighbors were $k\in\{15, 20, 30\}$,
and the best hyper-parameter setting option (the values of $w_1$, $w_2$, and $k$) was again chosen via internal 4-fold cross-validation. Since more is better when the number of iterations $m$ in Relief is concerned, this parameter was set to $m = |\dataset{}|$.

The possible numbers of labeled examples $L$ in the training datasets were $L\in\{50, 100, 200, 350, 500\}$ \cite{jurica:phd}.

\subsection{Evaluation pipeline}
\begin{figure}[!htb]
\centering
\includegraphics[width=.8\textwidth]{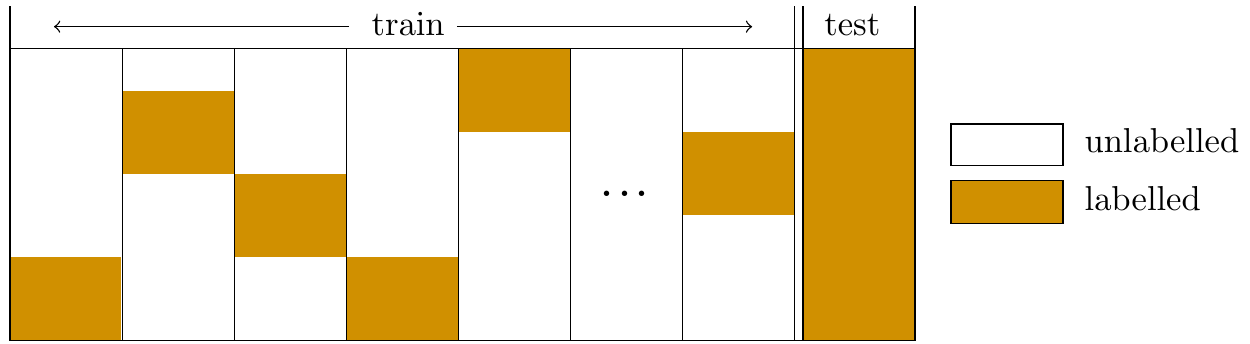}
\caption{Training and test set creation in SSL cross-validation: In the test fold, all examples keep their labels,
         whereas the folds that form the training set, together contain (approximately) $L$ labeled examples.}
\label{fig:xval}
\end{figure}

For the tasks of MLC and HMLC, the data come with predefined training and test parts ($\datatrain{}$ and $\datatest{}$). This is not the case for the tasks of 
classification, STR and MTR, therefore, 10-fold cross validation is performed. To obtain the training-test pairs in cross-validation,
we follow the procedure used by \cite{ssl-fr-stc}, as shown in Fig.~\ref{fig:xval}.

Each dataset $\dataset{}$ is randomly split into $x = 10$ folds which results in the test sets $\datatest{}_i$, $0\leq i < x$.
In contrast to cross-validation in the SL scenario, where $\datatrain{}_i = \cup_{j\neq i} \datatest{}_j$, we first define the copy $\datatest{}_i^L$ of $\datatest{}_i$ in which we keep the target values for $\lfloor L /(x - 1)\rfloor + r_i$ randomly selected examples
(orange parts of columns in Fig.~\ref{fig:xval}) and remove the others (white parts).
Here, $\lfloor \cdot \rfloor$ is the floor function, $r$ is the reminder of $L$ when divided by $x - 1$, and $r_i = 1$ if $i < r$ and $0$ otherwise. This assures that every training set $\datatrain{}_i^L = \cup_{j\neq i} \datatest{}_i^L$ contains a number of labeled examples as close as possible to $L$.

For the MLC and HMLC data, we can choose $L$ labeled instances from the training set and delete the target values for the others.
This is done for different numbers $L$ of labeled examples, and we make sure that the implication $L_1\leq L_2\Rightarrow $ \emph{labeled examples of $\datatrain{}_i^{L_1}$ are a subset of the labeled examples in $\datatrain{}_i^{L_2}$} holds.

The ranking evaluation proceeds as follows. First, SSL-ranking is computed from $\datatrain{}_i^L$ and its SL counterpart
is computed on the $\datatrain{}_i^L$ with the unlabeled examples removed.
Afterward, both rankings are evaluated on $\datatest{}_i^L$ (in the cases of MLC and HMLC, $\datatrain{}^L$ and $\datatest{}^L$ are used).

This is done by using the $k$NN algorithm with $k\in\{20, 40\}$ where weighted version of the standard squared Euclidean distance is used.
For two input vectors $\bm{x}^1$ in $\bm{x}^2$, the distance $d$ between them is defined as
$d(\bm{x}^1, \bm{x}^2) = \sum_{i = 1}^D w_i d_i^2(\bm{x}_i^1, \bm{x}_i^2)$,
where $d_i$ is defined as in Eq.~\eqref{eqn:metric}. The dimensional weights $w_i$ are defined as $w_i = \max\{\importance(x_i), 0\}$,
since Random Forest and Relief ranking can award a feature a negative score. In the degenerated case when the resulting values all equal $0$, we define $w_i = 1$, for all features $x_i$. The first step is necessary to ignore the features that are of lower importance than a randomly generated one would be.
The second step is necessary to ensure $d$ is well-defined. We chose more than one value of $k$ to show the qualitative differences
between the supervised and semi-supervised feature rankings.

The evaluation through $k$NN was chosen because of three main reasons. First it can be used for all the considered predictive modeling tasks. Second, this is a distance based method, hence, it can easily make use of the information 
contained in the feature importances in the learning phase. Third, $k$NN is simple: Its only parameter is the number of neighbors.
In the prediction stage, the neighbors' contributions to the predicted value are equally weighted, so we do not introduce additional parameters that would influence the performance.

\subsection{Evaluation measures}\label{sec:measures}

To asses the predictive performance of a $k$NN model, the following evaluation measures are used:
$F_1$ for classification (macro-averaged for multi-class problems),
Root Relative Squared Error (RRMSE) for STR and MTR, and area under the average precision-recall curve for MLC and HMLC (\pooled{}).
Their definitions are given in the Tab.~\ref{tab:measuers}.
In the cross-validation setting, we average the scores over the folds (taking test set sizes into account).

\begin{table}[ht]
    \centering
    \caption{Evaluation measures, for different predictive modeling tasks. The $F_1$ measure and \pooled{} are defined in terms of 
    precision $p = \tp{} / (\tp{} + \fp{})$ and recall $r = \tp{} / (\tp{} + \fn{})$,
    where the numbers $\tp{}$, $\fp{}$ and $\fn{}$ denote the number of true positive, false positive and false negative examples, respectively.}
    \label{tab:measuers}
    \begin{tabular}{l|l|l}
    \hline\noalign{\smallskip}
    tasks & measure & definition \\
    \noalign{\smallskip}\hline\noalign{\smallskip}
    classification &$F_1$ & $2/ (1/p + 1 / r)$\\
   MLC, HMLC &\pooled{} & area under the micro-averaged precision-recall curve \\
     STR, MTR & RRMSE & $\frac{1}{T}\sum_{j = 1}^T\sqrt{\frac{1}{|\datatest{}|}\sum_{(\bm{x}, \bm{y})\in\datatest{}} \frac{(\hat{\bm{y}}_j - \bm{y}_j)^2}{ \var{}(\datatest{}, \bm{y}_j)}}$  \\
    \noalign{\smallskip}\hline
    \end{tabular}
\end{table}

For each ranking and dataset,
we construct a curve that consist of points $(L, \mathit{performance}_L)$. The comparison of two methods is
then based either i) on these curves directly (see Fig.~\ref{fig:ssl-vs-sl}), or ii) on the area under the computed curves.

\subsection{The considered methods}\label{sec:considered-methods}
The methods that our proposed methods are compared to, depend on the predictive modeling task:
\begin{itemize}
    \item Classification: We have shown \cite{ssl-fr-stc} that ensemble-based ranking algorithms
have state-of-the-art performance. Thus, their and Relief's SSL and SL versions are compared against each other.
    \item STR: As mentioned before (Sec.~\ref{sec:related}), the existing SSL state-of-the-art competitor is Laplace,
    thus we compare Laplace, and the SL/SSL versions of both ensemble-based rankings and Relief-based rankings, against each other.
    \item MTR, MLC, HMLC: To the best of our knowledge, there are no existing methods that can perform feature ranking
    in the SSL structured output prediction scenarios, thus, we compare both versions of ensemble-based rankings and Relief-based rankings against each other.
\end{itemize}

Despite our best efforts, we could not obtain any existing implementation of the Laplace method, so we provide ours together with the rest of the code.
Also note that the ensemble-based and Relief-based methods work out of the box, i.e., no data preprocessing is necessary, whereas by design, Laplace can handle only numeric features. To overcome this issue, we extend the method by the following procedure: i) transform the nominal features using 1-hot encoding, ii) compute the Laplace scores $s_i$,
iii) for the originally nominal features $x_i$, define their score $s_i$ as the sum of the scores of the corresponding 1-hot encoded features, and, finally,
iv) define the importance scores $\importance{}_\text{Laplace}(x_i) = S + s - s_i$ (where $S$ and $s$ denote the maximum and the minimum of the scores, respectively).
The last step is necessary since less is better, for the originally computed Laplace scores. The transformation $s_i\mapsto S + s - s_i$ maps $S$ to $s$ and vice-versa,
thus, the scale remains intact.
The other problem of the method are constant features (they cause $0/0$ values), present, for example, in QSAR data: These had to be manually removed.

\section{Results}\label{sec:results}

Unless stated otherwise, the rankings are compared in terms of the areas under the performance curves (see Sec.~\ref{sec:measures}).
When a SSL-ranking is compared to a SL-ranking, and the difference $\Delta$ between the two performances is computed,
$\Delta > 0$ always corresponds to the case when the SSL-ranking performs better.

\subsection{The optimal ensemble method for ensemble-based ranking}\label{sec:best-ens}
We first determine the most appropriate ensemble method, for each of the three ensemble scores,
and their two versions (SSL and SL). The results in Tab.~\ref{tab:best-ensemble} give the average ranks of the ensemble methods in each setting,
in terms of the areas under the performance curves.

\begin{table}[htbp]
  \centering
  \caption{Average ranks of the considered SSL and SL ensembles, for a fixed ensemble-based score and predictive modeling task.
  The best ranks are shown in bold, unless all three methods perform equally well. In the case of ties, we bold the most efficient method (see Tab.~\ref{tab:ens-time-ranks}).}
  \label{tab:best-ensemble}
    \begin{tabular}{|c|l|ccl|ccl|}
    \hline
    \multirow{2}[2]{*}{task} & \multicolumn{1}{c|}{\multirow{2}[2]{*}{score}} & \multicolumn{3}{c|}{SSL ensemble} & \multicolumn{3}{c|}{SL ensemble} \\
          &       & RFs & ETs &  bagging & RFs & ETs & bagging \\
    \hline
    \multirow{3}[2]{*}{classification} & Genie3 & 2.00  & 2.15  & \textbf{1.85}  & \textbf{1.69}  & 2.46  & 1.85 \\
          & Random Forest & \textbf{1.92}  & 2.15  & 1.92  & 2.00  & 2.08  & \textbf{1.92} \\
          & Symbolic & \textbf{1.77}  & 2.23  & 2.00  & \textbf{1.77}  & 2.23  & 2.00 \\
    \hline
    \multirow{3}[2]{*}{MLC} & Genie3 & \textbf{1.50}  & 2.67  & 1.83  & 2.17  & 2.17  & \textbf{1.67} \\
          & Random Forest & \textbf{2.00}  & 2.00  & 2.00  & \textbf{1.67}  & 2.00  & 2.33 \\
          & Symbolic & \textbf{1.33}  & 2.33  & 2.33  & 2.00  & 2.50  & \textbf{1.50} \\
    \hline
    \multirow{3}[2]{*}{HMLC} & Genie3 & \textbf{1.67}  & 2.00  & 2.33  & \textbf{1.67}  & 1.83  & 2.50 \\
          & Random Forest & 2.17  & \textbf{1.83}  & 2.00  & 1.83  & \textbf{1.67}  & 2.50 \\
          & Symbolic & 2.17  & 2.00  & \textbf{1.83}  & \textbf{1.67}  & 2.00  & 2.33 \\
    \hline
    \multirow{3}[2]{*}{STR} & Genie3 & \textbf{1.67}  & 2.00  & 2.33  & 2.17  & 2.00  & \textbf{1.83} \\
          & Random Forest & 2.00  & \textbf{1.83}  & 2.17  & 2.67  & \textbf{1.67}  & 1.67 \\
          & Symbolic & 2.17  & \textbf{1.50}  & 2.33  & 2.33  & \textbf{1.83}  & 1.83 \\
    \hline
    \multirow{3}[2]{*}{MTR} & Genie3 & 2.14  & \textbf{1.86}  & 2.00  & 2.43  &\textbf{1.71}  & 1.86 \\
          & Random Forest & 2.29  & 2.14  & \textbf{1.57}  & 2.43  & \textbf{1.71}  & 1.86 \\
          & Symbolic & \textbf{2.00}  & 2.00  & 2.00  & 2.29  & \textbf{1.71}  & 2.00 \\
    \hline
    \end{tabular}
\end{table}

We observe that for both regression tasks (STR and MTR), RFs ensembles almost never perform best (with the exception of Genie3 SSL-rankings),
whereas for the other three classification-like tasks, they quite consistently outperform the other two ensemble methods.
The differences among the average ranks are typically not considerable (with the exception of the most of the MLC rankings, and supervised MTR rankings)
which is probably due to the fact that the split selection mechanisms of the considered ensemble methods are still quite similar,
and the trees are fully-grown, so sooner or later, a relevant feature appears in the node. In the case of ties, we choose the
more efficient one (see Tab.~\ref{tab:ens-time-ranks}): RFs are always the most efficient, whereas the second place is determined by the number of possible splits
per feature. For lower values (e.g., when most of the features are binary, as is the case in MLC and HMLC data), bagging is faster than ETs.
\begin{table}[htbp]
  \centering
  \caption{Average ranks of the ensemble methods, in terms of induction times.}\label{tab:ens-time-ranks}%
    \begin{tabular}{|l|rrr|}
    \hline
    task  & \multicolumn{1}{l}{RFs} & \multicolumn{1}{l}{ETs} & \multicolumn{1}{l|}{bagging} \\
    \hline
    classification & \textbf{1.00}  & 2.23  & 2.77\\
    MLC   & \textbf{1.00}  & 2.67  & 2.33 \\
    HMLC  & \textbf{1.00}  & 2.50  & 2.50 \\
    STR   & \textbf{1.17}  & 2.33  & 2.50 \\
    MTR   & \textbf{1.29}  & 1.71  & 3.00\\
    \hline
    \end{tabular}%
\end{table}%

To make the later graphs more readable, we plot, for every score, only the curve that corresponds to the most suitable ensemble method for this score.

\subsection{Qualitative difference between SSL and SL rankings}
\begin{figure*}[h!]
\centering
\begingroup
    \captionsetup[subfigure]{width=0.49\textwidth}
    \subfloat[classification: digits, $20$NN\label{fig:ssl-vs-sl:stc:20}]{
    \includegraphics[trim={0.68cm 0.7cm 0.72cm 0.7cm},clip,width=0.47\textwidth]{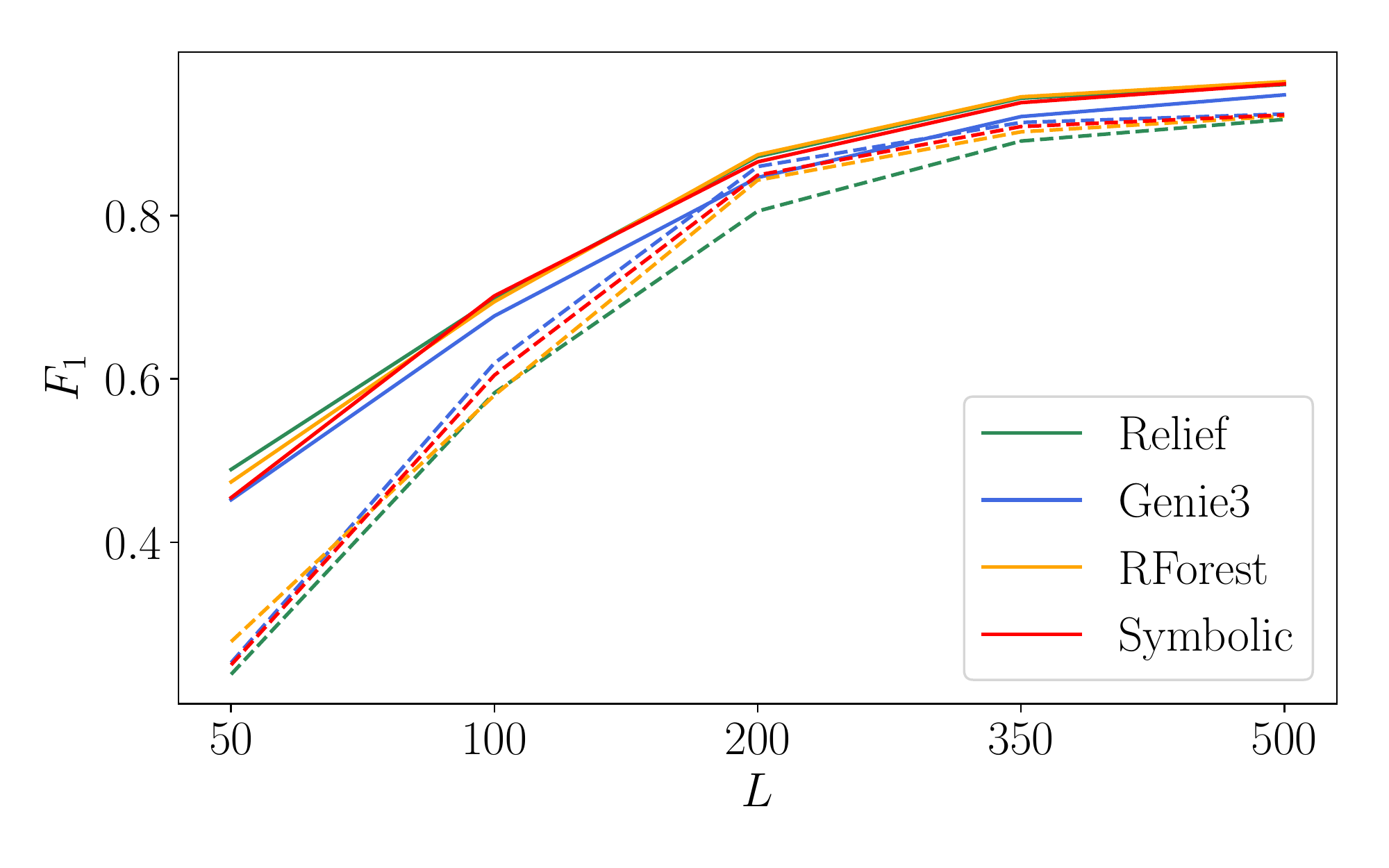}}
\endgroup
\begingroup
    \captionsetup[subfigure]{width=0.49\textwidth}
    \subfloat[classification: digits, $40$NN\label{fig:ssl-vs-sl:stc:40}]{
    \includegraphics[trim={0.68cm 0.7cm 0.72cm 0.7cm},clip,width=0.47\textwidth]{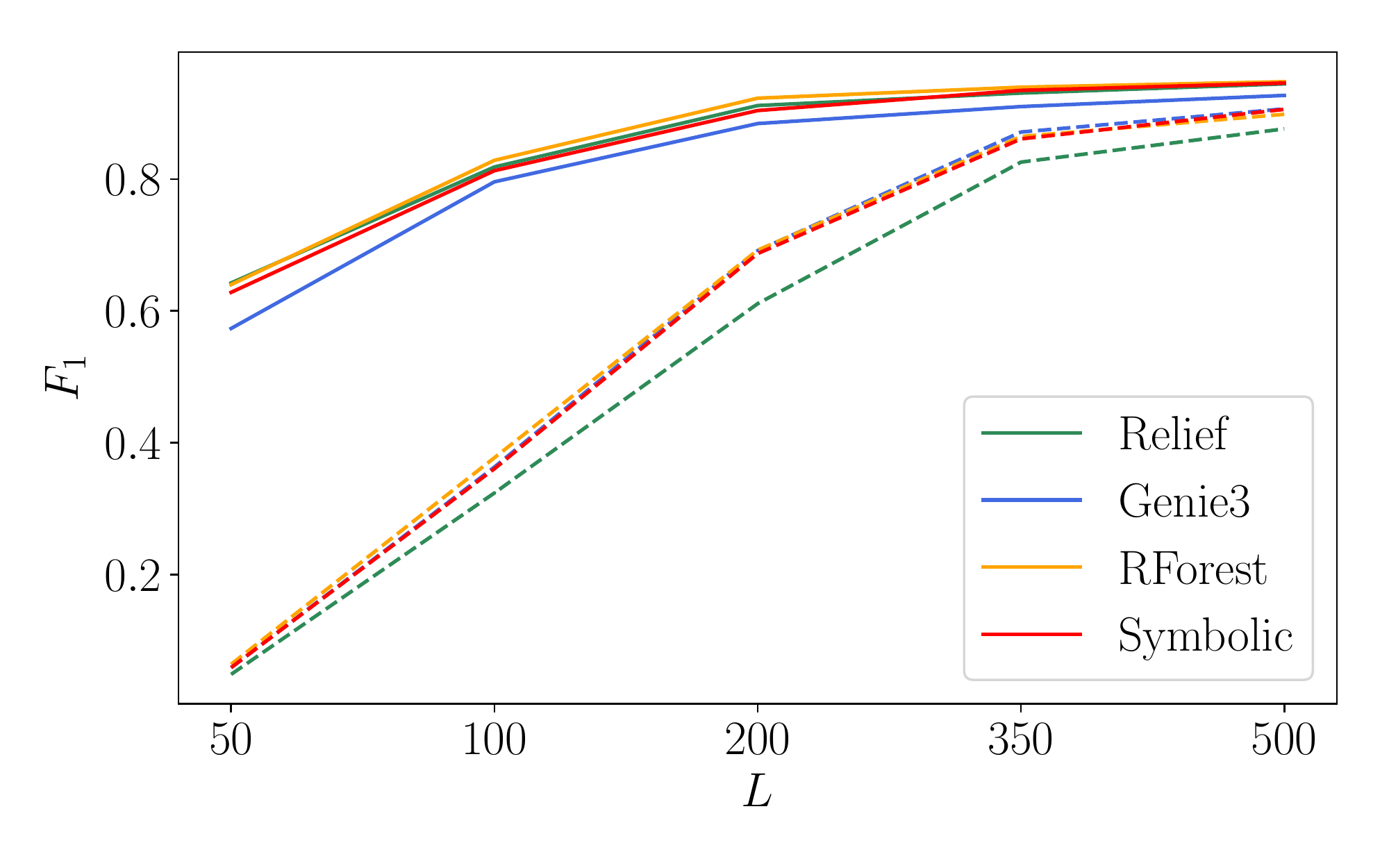}}
\endgroup

\begingroup
    \captionsetup[subfigure]{width=0.49\textwidth}
    \subfloat[MLC: genbase, $20$NN\label{fig:ssl-vs-sl:mlc:20}]{
    \includegraphics[trim={0.7cm 0.7cm 0.7cm 0.7cm},clip,width=0.47\textwidth]{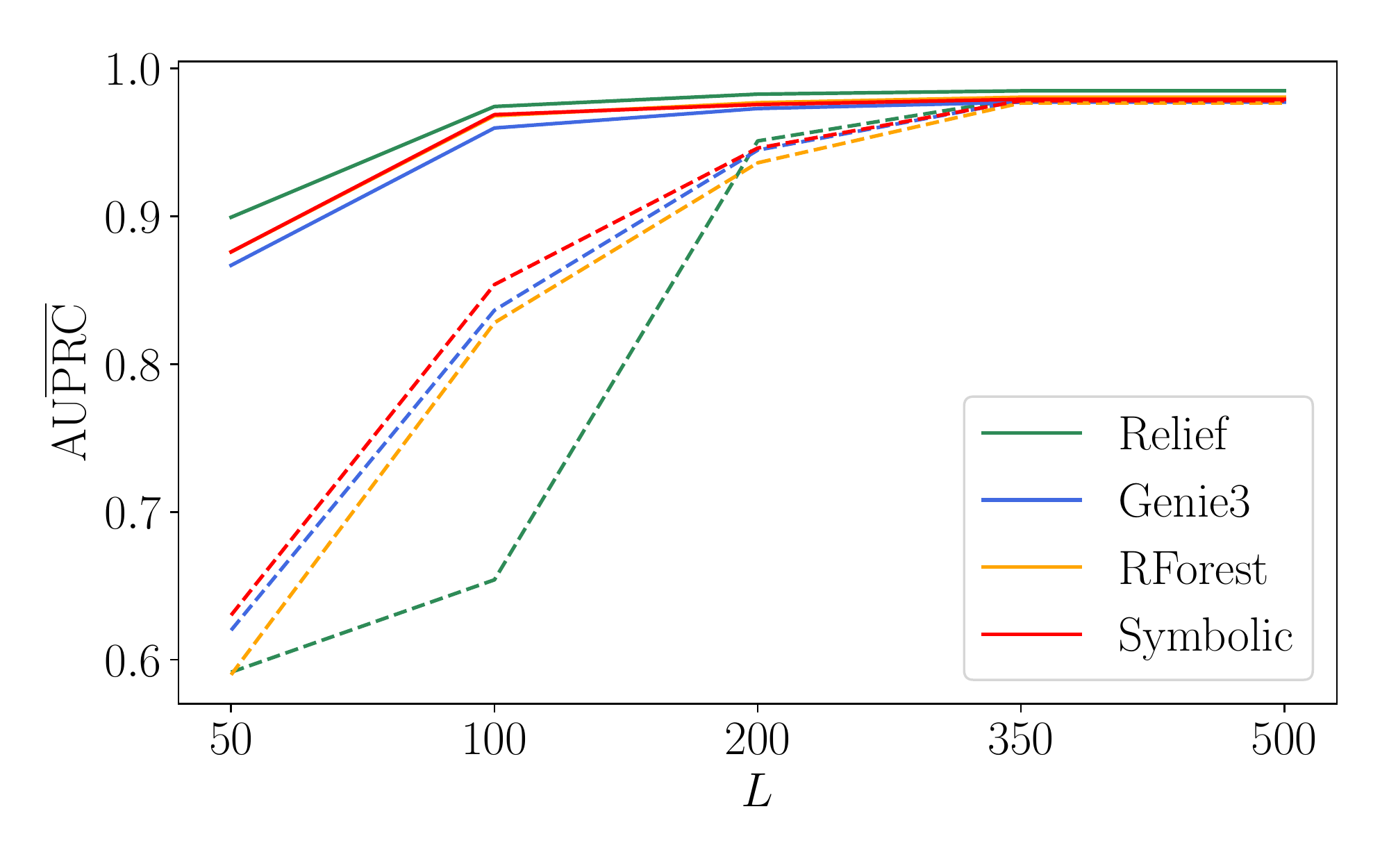}}
\endgroup
\begingroup
    \captionsetup[subfigure]{width=0.49\textwidth}
    \subfloat[MLC: genbase, $40$NN\label{fig:ssl-vs-sl:mlc:40}]{
    \includegraphics[trim={0.7cm 0.7cm 0.7cm 0.7cm},clip,width=0.47\textwidth]{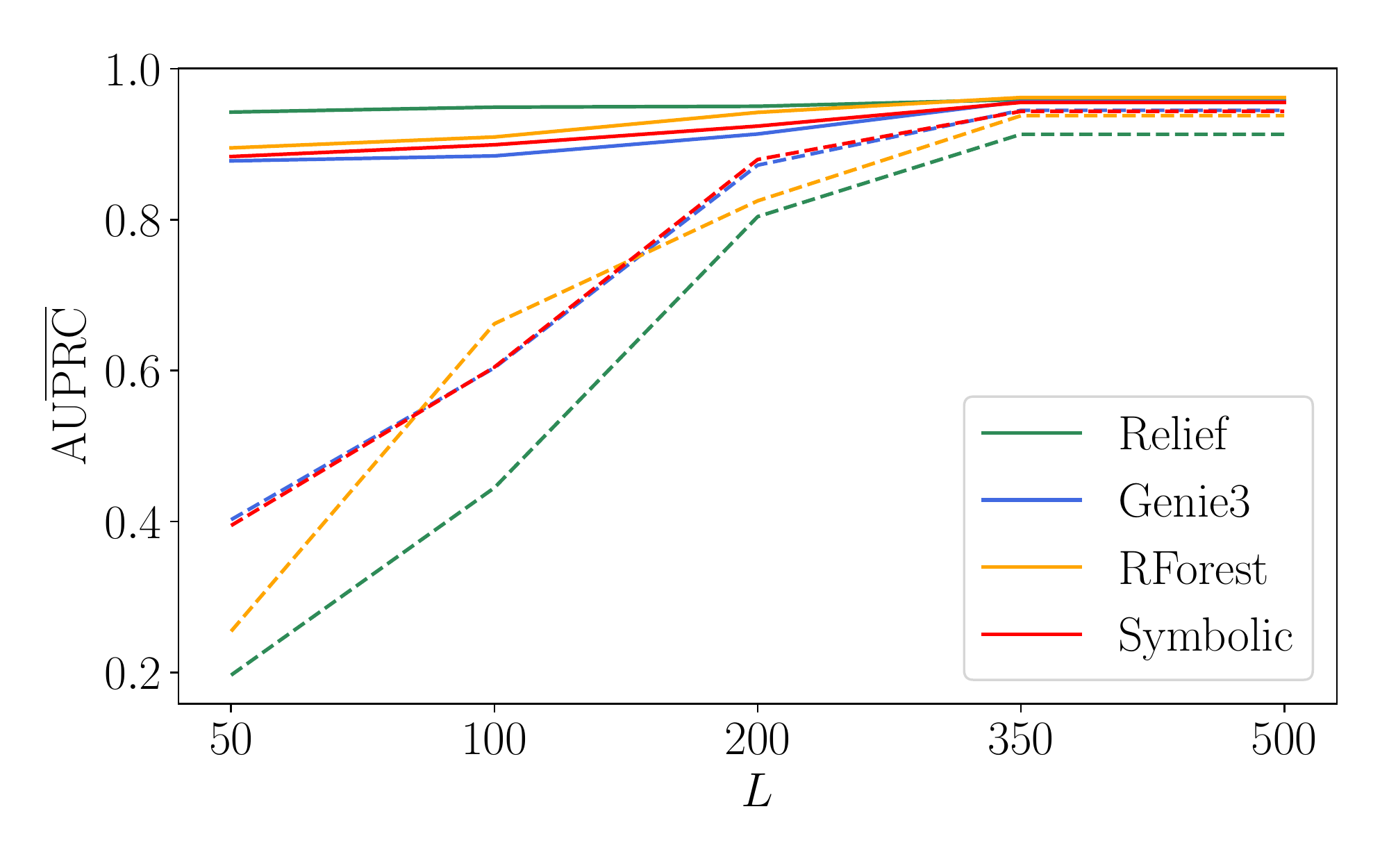}}
\endgroup

\begingroup
    \captionsetup[subfigure]{width=0.49\textwidth}
    \subfloat[HMLC: ecogen, $20$NN\label{fig:ssl-vs-sl:hmlc:20}]{
    \includegraphics[trim={0.7cm 0.7cm 0.7cm 0.7cm},clip,width=0.47\textwidth]{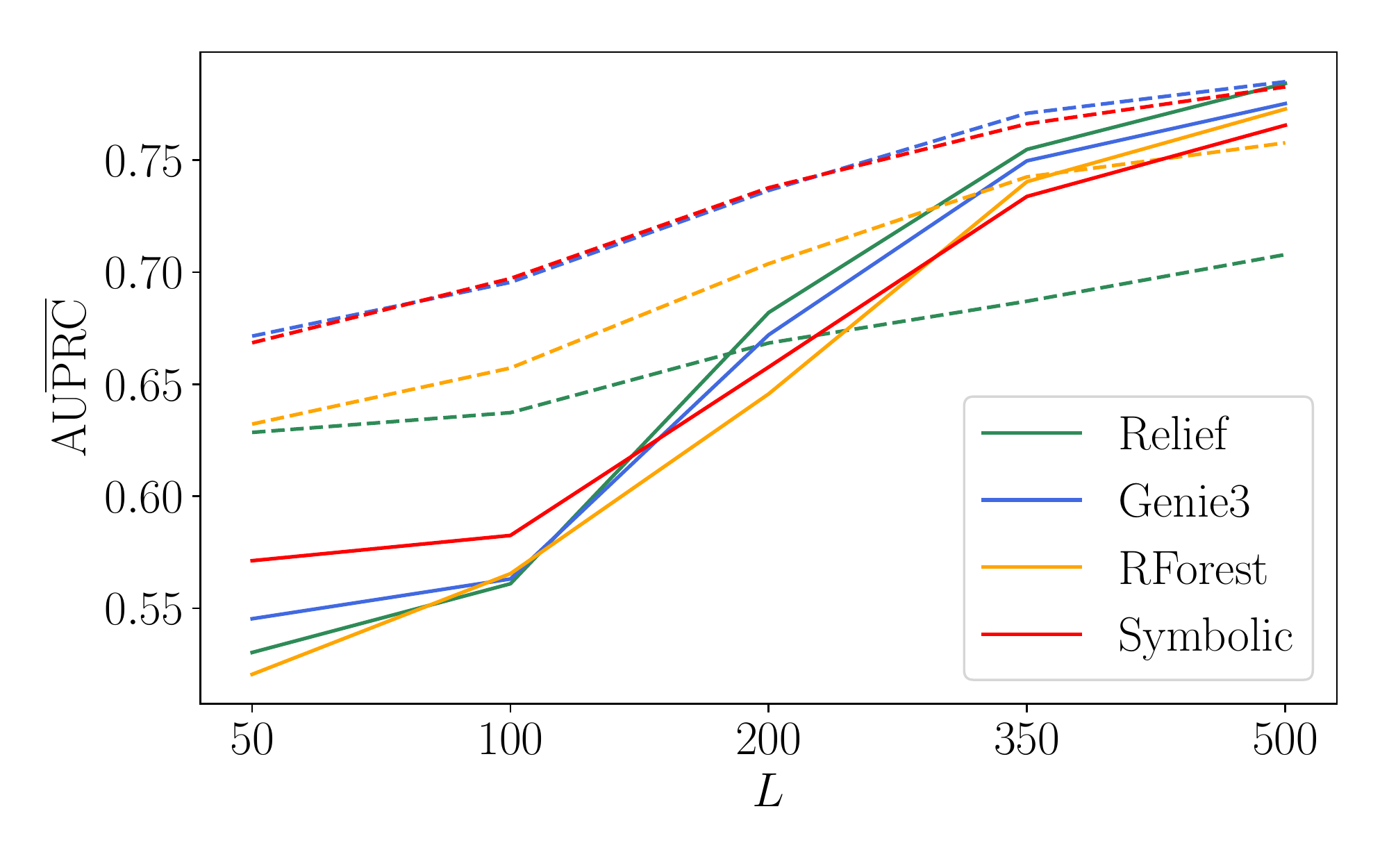}}
\endgroup
\begingroup
    \captionsetup[subfigure]{width=0.49\textwidth}
    \subfloat[HMLC: ecogen, $40$NN\label{fig:ssl-vs-sl:hmlc:40}]{
    \includegraphics[trim={0.7cm 0.7cm 0.7cm 0.7cm},clip,width=0.47\textwidth]{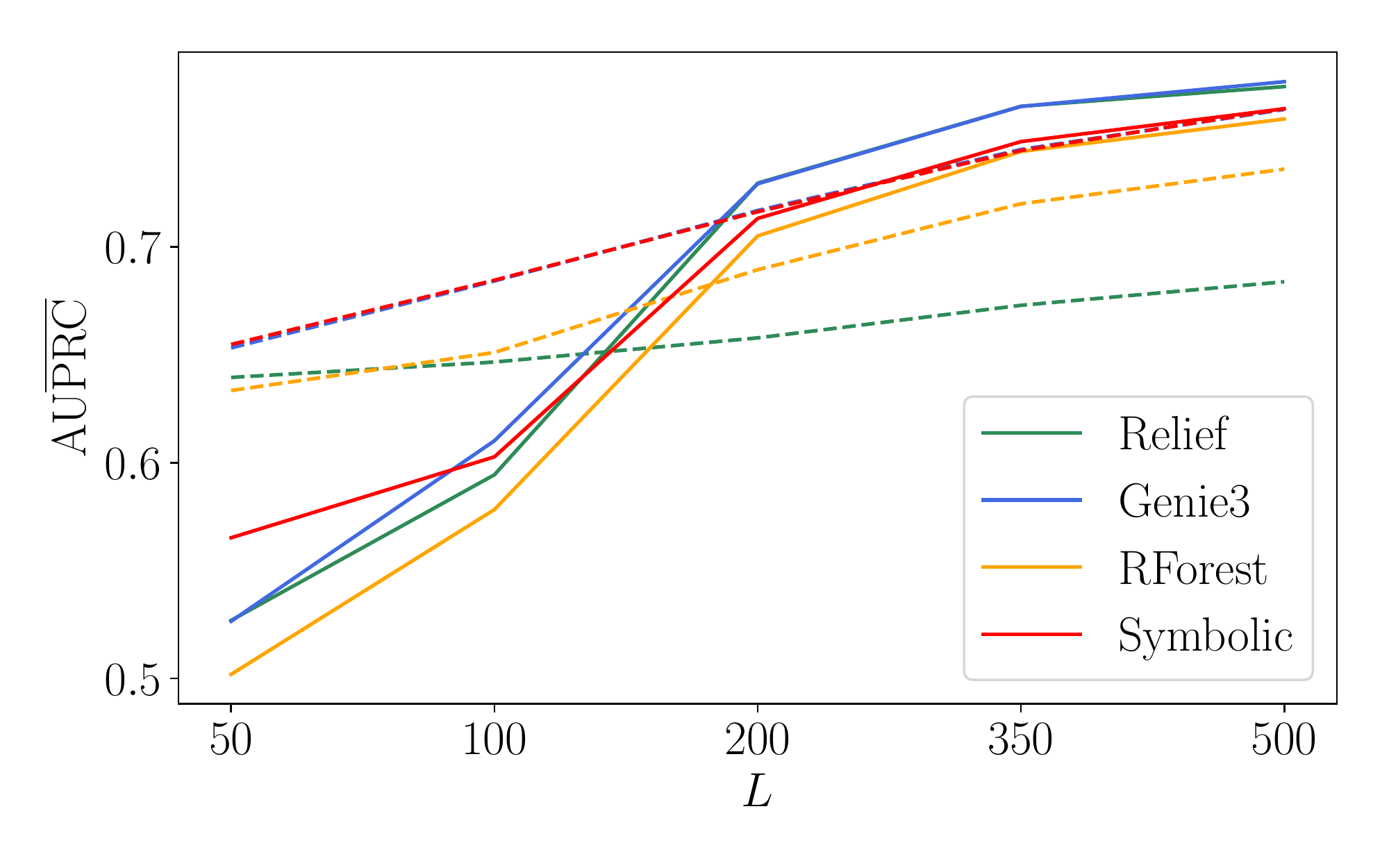}}
\endgroup

\begingroup
    \captionsetup[subfigure]{width=0.49\textwidth}
    \subfloat[STR: treasury, $20$NN\label{fig:ssl-vs-sl:str:20}]{
    \includegraphics[trim={0.7cm 0.7cm 0.7cm 0.7cm},clip,width=0.47\textwidth]{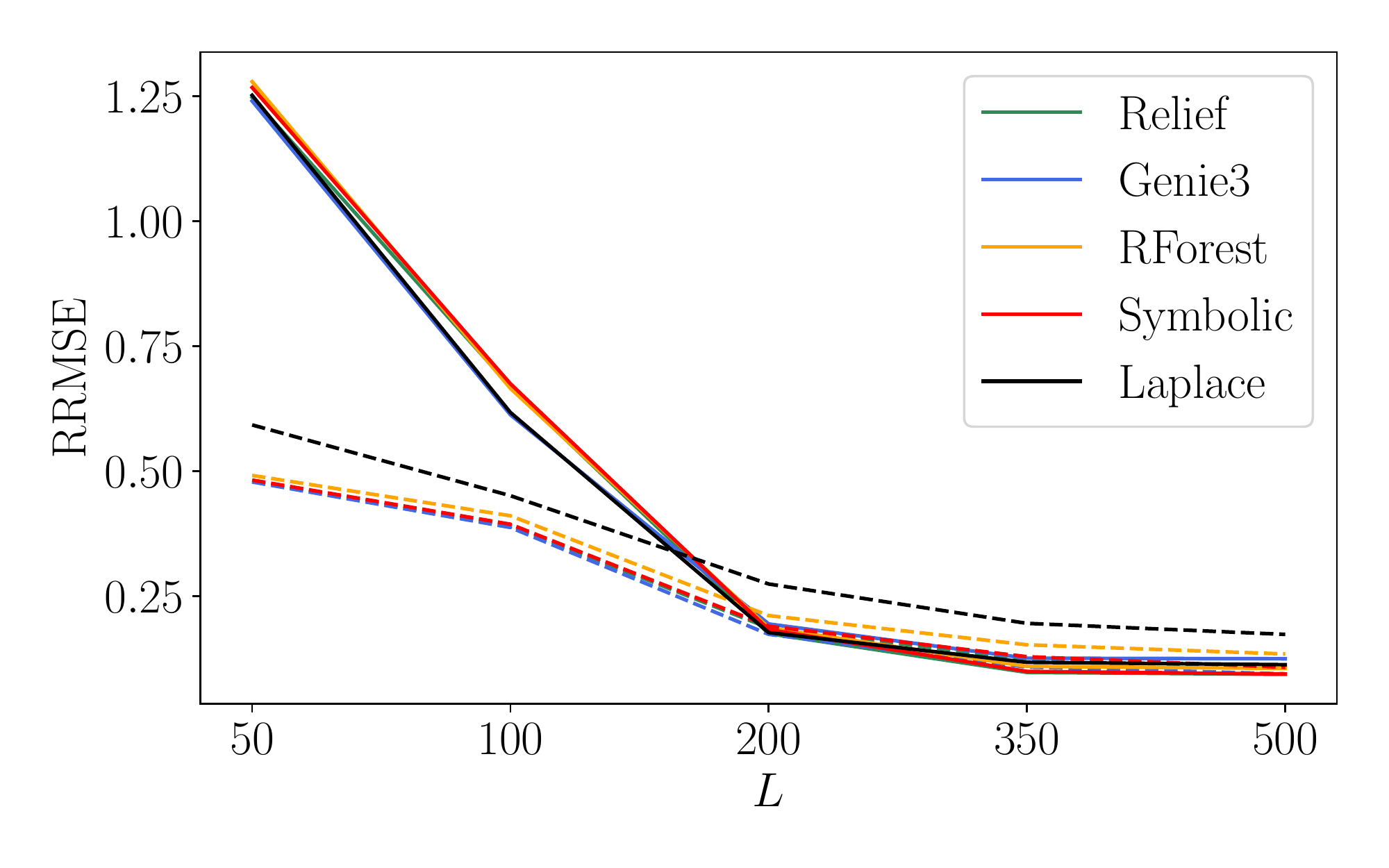}}
\endgroup
\begingroup
    \captionsetup[subfigure]{width=0.49\textwidth}
    \subfloat[STR: treasury, $40$NN\label{fig:ssl-vs-sl:str:40}]{
    \includegraphics[trim={0.7cm 0.7cm 0.7cm 0.7cm},clip,width=0.47\textwidth]{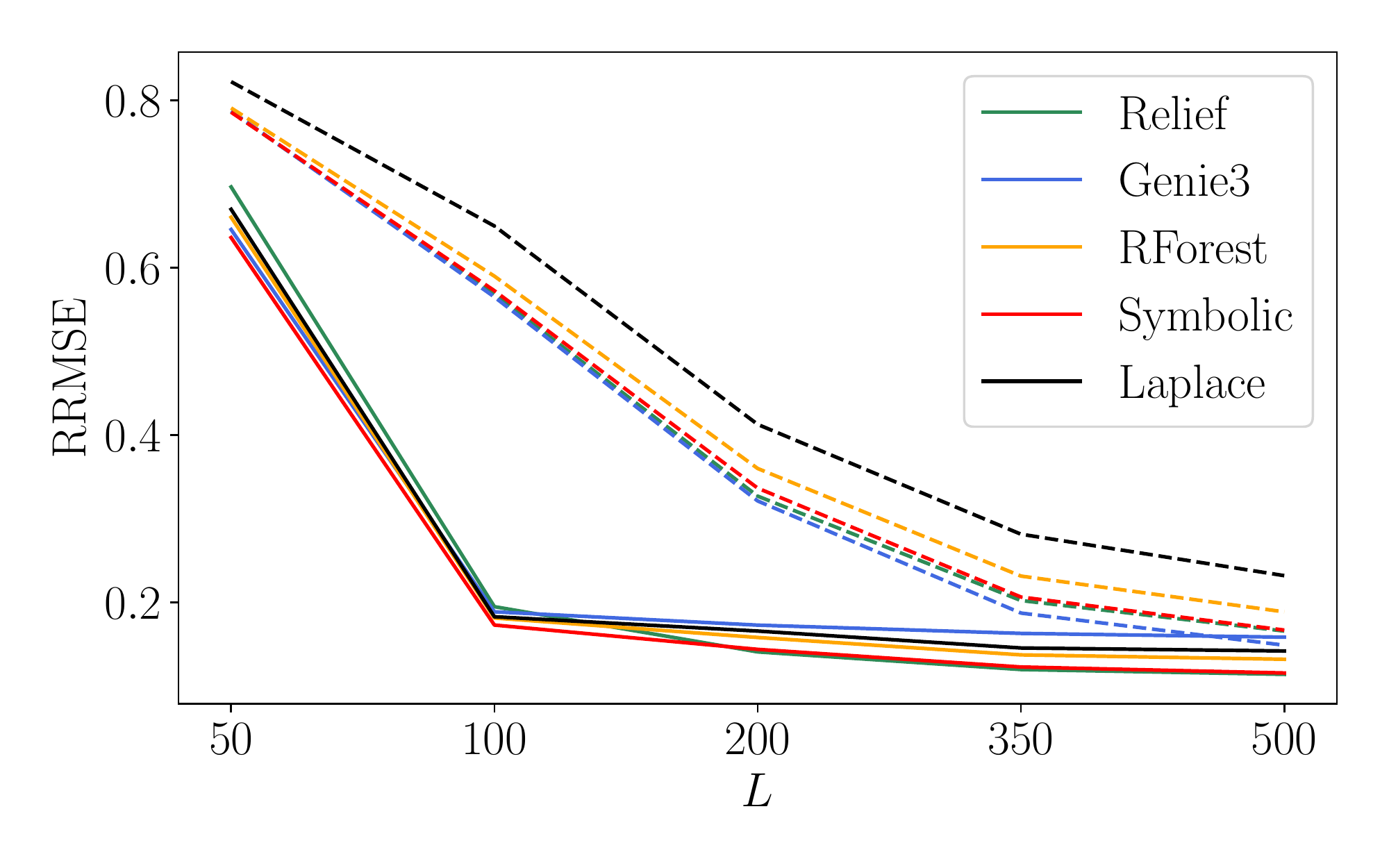}}
\endgroup

\begingroup
    \captionsetup[subfigure]{width=0.49\textwidth}
    \subfloat[MTR: oes10, $20$NN\label{fig:ssl-vs-sl:mtr:20}]{
    \includegraphics[trim={0.7cm 0.7cm 0.7cm 0.7cm},clip,width=0.47\textwidth]{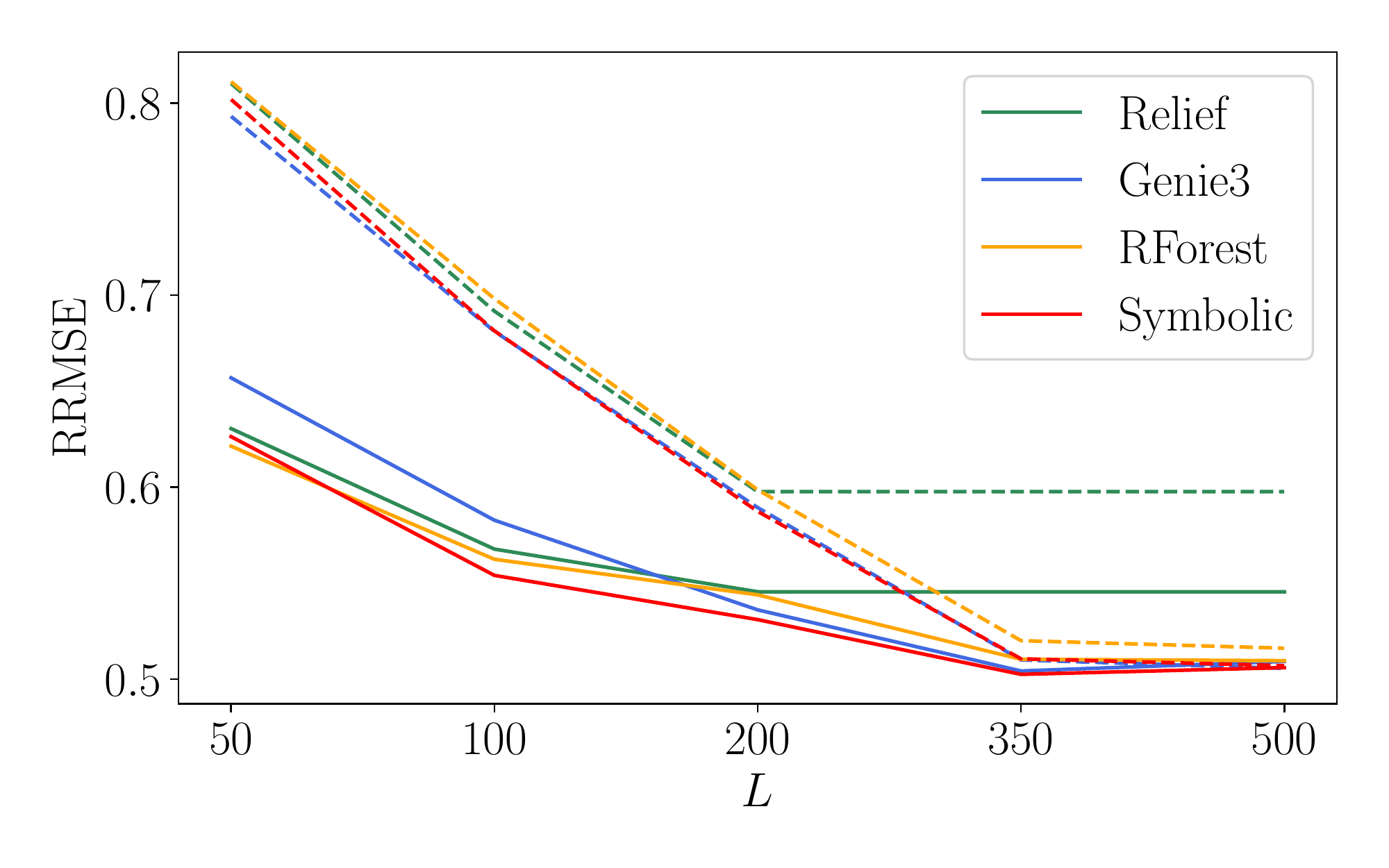}}
\endgroup
\begingroup
    \captionsetup[subfigure]{width=0.49\textwidth}
    \subfloat[MTR: oes10, $40$NN\label{fig:ssl-vs-sl:mtr:40}]{
    \includegraphics[trim={0.7cm 0.7cm 0.7cm 0.7cm},clip,width=0.47\textwidth]{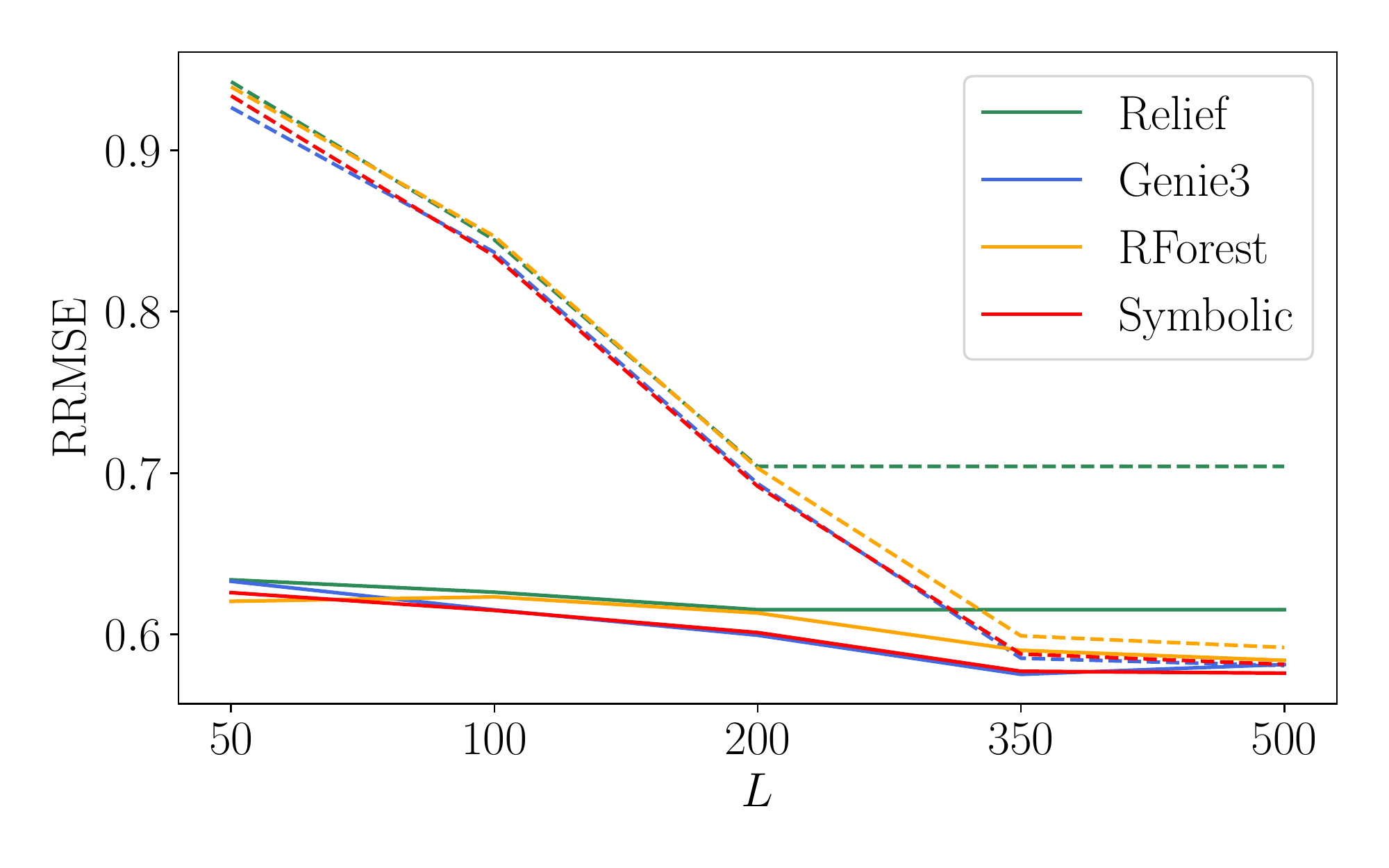}}
\endgroup
\caption{Comparison of the SL and SSL feature rankings, for different predictive modeling tasks.
The curves for the SSL and the SL versions of a ranking are shown as a solid and a dashed line of the same color.
The graphs in the left column use $20$NN models in the  evaluation, whereas those in the right, use $40$NN models.}
\label{fig:ssl-vs-sl}
\end{figure*}

We first discuss the qualitative difference between the SSL-rankings and their supervised counterparts.
In the process of obtaining a feature ranking, the SSL-version of the ranking algorithm sees more examples than its supervised version,
and it turns out that this is well-reflected in the results. Fig.~\ref{fig:ssl-vs-sl} shows the results for five datasets (one dataset, for each task)
and the performance of the rankings, as assessed by $k$NN models, for $k\in\{20, 40\}$. Those two values of $k$ are used to show that SSL-rankings
tend to capture a more global picture of data, whereas the supervised ones reflect a more local one.
\begin{table}[htb]
    \centering
    \caption{Proportions of the computed feature rankings whose SSL-version captures more global properties of the data, as compared to its supervised version.
    The differences $\delta_{20}$ and $\delta_{40}$ of the areas under the performance curves of $20$NN and $40$NN models are computed
    (always in a way that $\delta > 0$ means that SSL-version performs better). Therefore, if $\Delta = \delta_{40} -\delta_{20} > 0$,
    then the SSL-version of the ranking is more global, and is more local if $\Delta < 0$.}
    \label{tab:locality}
    \begin{tabular}{|l|ccccc|}
        \hline
        task & classification & MLC & HMLC & STR & MTR \\
        \hline
        $P[\Delta > 0]$ & 0.73 & 0.83 & 0.96 &  1.00 & 0.93 \\
        \hline
    \end{tabular}
\end{table}
This phenomenon is most visible in the two regression datasets. In the case of the \texttt{treasury} dataset,
SSL-rankings perform worse than supervised ones on the local scale for smaller numbers $L$ of labeled examples (Fig.~\ref{fig:ssl-vs-sl:str:20}),
and are equal or better for $L\geq 200$. However, on the global scale (Fig.~\ref{fig:ssl-vs-sl:str:40}), the SSL-rankings are clear winners.
A similar situation is observed for the other datasets in Fig.~\ref{fig:ssl-vs-sl}, and also in general.

Tab.~\ref{tab:locality} reveals that for the vast majority of the rankings (and datasets), the SSL rankings are more global.
This proportion is the highest for STR data (it even equals $100\%$), and is understandably the lowest for classification, where the datasets
have the smallest number of examples on average.

\FloatBarrier

\subsection{Can unlabeled data improve feature rankings?}
To answer this question, we compare the SSL versions of the proposed feature rankings to their supervised counterparts.
In the previous section, we explained why sometimes the answer is not straightforward and depends on whether one is interested in a global or local scale.
Since the question is whether the ranking can be improved by using unlabeled data, and given the qualitative differences between the SSL- and SL-versions
of the rankings from the previous section, we fix the number of neighbors to $k = 40$.

We start with the classification results given in Tab.~\ref{tab:classification}.
\begin{table}[htbp]
  \centering
  \caption{The differences $\Delta$ of areas under the curves of $F_1$-values of the $40$NN models with distance weights based on SSL-rankings and SL-rankings.}\label{tab:classification}%
    \begin{tabular}{|c|l|rrrr|r|}
    \hline
    &datasets & \multicolumn{1}{l}{Genie3} & \multicolumn{1}{l}{RForest} & \multicolumn{1}{l}{Symbolic} & \multicolumn{1}{l|}{Relief} & CH \\
    \hline
    \multirow{13}{*}{\STAB{\rotatebox[origin=c]{90}{classification}}} &
    Arrhythmia & 0.039 & 0.008 & 0.022 & 0.006 & 0.02\\
    &Bank       & 0.064 & 0.067 & 0.061 & 0.050 & -0.00\\
    &Chess      & -0.084& 0.021 & -0.081& 0.022 & 0.22\\
    &Dis        & 0.066 & 0.046 & 0.050 & 0.123 & 0.00\\
    &Gasdrift   & 0.041 & 0.038 & 0.053 & 0.109 & 0.02\\
    &Pageblocks & 0.272 & 0.250 & 0.250 & 0.243 & 0.03\\
    &Phishing   & -0.125& -0.128& -0.132& -0.115& -0.00\\
    &Tic-tac-toe& 0.148 & 0.225 & 0.152 & 0.141 & 0.70\\
    &Aapc       & 0.115 & 0.041 & 0.067 & 0.110 & 0.34\\
    &Coil2000   & 0.019 & 0.029 & 0.022 & 0.020 & -0.00\\
    &Digits     & 0.170 & 0.204 & 0.198 & 0.245 & -0.00\\
    &Pgp        & 0.043 & 0.113 & 0.096 & 0.139 & 0.00\\
    &Thyroid    & 0.288 & 0.268 & 0.285 & 0.265 & 0.01\\
    \hline
    \end{tabular}%
\end{table}%

From the mainly positive numbers in the table, one can conclude that SSL-rankings successfully recognize the structure of data,
and outperform their supervised analogs, even in most of the cases where the CH values are low, e.g., for \texttt{digits} dataset in Fig.~\ref{fig:ssl-vs-sl:stc:20},
or, most notably, for \texttt{pageblocs}.
\begin{table}[htbp]
  \centering
  \caption{The differences $\Delta$ of areas under the curves of \pooled{}-values of the $40$NN models whose distance weights base on SSL-ranking and SL-ranking.} \label{tab:mlc}%
    \begin{tabular}{|c|l|rrrr|c|}
    \hline
    &datasets & \multicolumn{1}{l}{Genie3} & \multicolumn{1}{l}{RForest} & \multicolumn{1}{l}{Symbolic} & \multicolumn{1}{l|}{Relief} & CH\\
    \hline
    \multirow{6}{*}{\STAB{\rotatebox[origin=c]{90}{MLC}}} &
    Bibtex            & -0.115& -0.078& -0.100& -0.100& 0.02   \\
    &Birds & 0.039 & 0.052 & 0.021 & 0.029 & 0.05   \\
    &Emotions          & 0.012 & 0.028 & 0.011 & 0.044 & 0.04   \\
    &Genbase            & 0.091 & 0.121 & 0.094 & 0.189 & 0.26   \\
    &Medical           & -0.067& 0.008 & -0.058& 0.014 & 0.04   \\
    &Scene             & 0.045 & 0.048 & 0.063 & 0.119 & 0.21   \\
    \hline   
    \multirow{6}{*}{\STAB{\rotatebox[origin=c]{90}{HMLC}}} &
    Clef07a-is       & -0.097& -0.066& -0.102& -0.041& 0.05 \\  
    &Ecogen             & -0.007& -0.003& -0.018& 0.051 & 0.03 \\  
    &Enron-corr      & -0.068& -0.062& -0.023& -0.064& 0.03 \\  
    &Expr-yeast-fun  & -0.090& -0.103& -0.086& -0.071& 0.00 \\  
    &Gasch1-yeast-FUN& -0.080& -0.087& -0.084& -0.096& 0.01 \\  
    &Pheno-yeast-FUN & -0.032& -0.031& -0.036& -0.029& 0.00 \\  
    \hline
    \end{tabular}%
\end{table}%

Continuing with the results for MLC (the upper part of Tab.~\ref{tab:mlc}), we first see that CH values are rather low,
since, in contrast to the ARI values from classification, correction for chance is not incorporated into these CH values.
An exception to this are the \texttt{genbase} (see Figs.~\ref{fig:ssl-vs-sl:mlc:20} and \ref{fig:ssl-vs-sl:mlc:40}) and the \texttt{scene} dataset.
For both datasets, the SSL-versions of the rankings outperform their SL-analogs. This also holds for the \texttt{birds} and \texttt{emotions}
datasets, for all rankings, and additionally for the \texttt{medical} dataset in the case of Relief.

The bottom part of Tab.~\ref{tab:mlc} gives the results for HMLC datasets. One can notice that Asm.~\ref{lab:ch} is never satisfied (low CH values),
and that SSL-scores mostly could not overcome this, with the exception of Relief rankings on the \texttt{ecogen} dataset.
However, inspecting the corresponding curves in detail (Fig.~\ref{fig:ssl-vs-sl:hmlc:40}), reveals that the negative differences
in the performance of SSL-rankings and SL-rankings are mostly due to the bad start of SSL-rankings: For $L\geq 200$, the SSL-versions prevail.

\begin{table}[htbp]
  \centering
  \caption{The differences $\Delta$ of areas under the curves of RRMSE-values of the $40$NN models with distance weights based on SSL-rankings and SL-rankings.}\label{tab:r}
    \begin{tabular}{|c|l|rrrrr|c|}
    \hline
    &datasets & \multicolumn{1}{l}{Genie3} & \multicolumn{1}{l}{RForest} & \multicolumn{1}{l}{Symbolic} & Relief & \multicolumn{1}{l|}{Laplace} & CH \\
    \hline
    \multirow{6}{*}{\STAB{\rotatebox[origin=c]{90}{STR}}} &
    CHEMBL2850          & -0.047& -0.063& 0.014 &-0.092& -0.010&  0.09 \\ 
    &CHEMBL2973          & -0.143& -0.103& -0.114&-0.168& -0.109&  0.18 \\ 
    &Mortgage            & 0.074 & 0.092 & 0.097 &0.078 & 0.120 &  0.57 \\ 
    &Pol                 & 0.027 & 0.249 & 0.127 &-0.049& 0.278 &  0.12 \\ 
    &QSAR                & -0.347& -0.446& -0.442&-0.523& -0.262&  0.20 \\ 
    &Treasury            & 0.118 & 0.172 & 0.165 &0.155 & 0.215 &  0.54 \\ 
    \hline                                       
    \multirow{7}{*}{\STAB{\rotatebox[origin=c]{90}{MTR}}} &
    Atp1d               & 0.048 & 0.024 & 0.048 &0.093 &       &  0.49 \\ 
    &CollembolaV2        & -0.048& -0.014& -0.010&-0.002&       &  0.02 \\ 
    &Edm1                & 0.002 & 0.018 & 0.004 &0.006 &       &  0.23 \\ 
    &Forestry-LIDAR-IRS& -0.115& -0.070& -0.101&-0.114&       &  0.19 \\ 
    &Oes10               & 0.083 & 0.084 & 0.083 &0.122 &       &  0.63 \\ 
    &Scm20d              & -2.357& -2.317& -2.295&-2.281&       &  0.16 \\ 
    &Soil-quality       & -0.044& -0.085& -0.080&-0.111&       &  0.07 \\
    \hline
    \end{tabular}%
\end{table}%

We finish this section with the regression results. The upper part of Tab.~\ref{tab:r} shows that when CH is well-setisfied,
i.e., for the datasets \texttt{mortgage} and \texttt{treasury} (see Fig.~\ref{fig:ssl-vs-sl:str:40}), the SSL-rankings outperform the SL-rankings.
Moreover, this also holds for the \texttt{pol} data (except for the Relief rankings). Inspecting the datasets where negative values are present
(most notably the \texttt{qsar} dataset) reveals the same phenomenon as in HMLC case: for extremely low values of $L$, e.g., $L = 50$,
the SSL-rankings do not perform well, possibly because knowing the labels of $50$ out of approximately $2000$ examples simply does not suffice.
With more and more labels known, the performance of SSL-rankings drastically improves, while the performance of SL-rankings stagnates.
Finally, for $L\geq 200$ or $L\geq 350$, all SSL-rankings again outperform the SL-ones.

Similar findings hold for the MTR data and the results in the bottom part of Tab.~\ref{tab:r}. The SSL-rankings perform well from the very beginning
on the three datasets where CH holds the most, i.e., \texttt{oes10} (see Fig.~\ref{fig:ssl-vs-sl:mtr:40}), \texttt{atp1d}, and \texttt{edm1},
but can only catch up with the SL-rankings (and possibly outperform them) for larger values of $L$ in the other cases.

\subsection{Which SSL-ranking performs best?}
To answer this question, we compare the predictive performances of the corresponding $40$NN models and report their ranks in Tab.~\ref{tab:ranking-quality}.
The results reveal that, for majority of the tasks, ensemble-based rankings perform best, however, in some cases, the winners are not clear, e.g., in the case of the classification. Still, Symbolic ranking quite clearly outperforms the others on both regression tasks, STR and MTR.

To complement this analysis, we also compute the average ranks of the algorithms for their induction times.
\begin{table}[htbp]
  \centering
  \caption{The average ranks of different SSL-ranking algorithms that base on the performance of the
  corresponding $40$NN models.}\label{tab:ranking-quality}%
    \begin{tabular}{|l|rrrrr|}
    \hline
    task  & \multicolumn{1}{l}{Genie3} & \multicolumn{1}{l}{Random Forest} & \multicolumn{1}{l}{Symbolic} & \multicolumn{1}{l}{Relief} & \multicolumn{1}{l|}{Laplace} \\
    \hline
    classification & 2.62  & 2.46  & 2.62  & \textbf{2.31}  &  \\
    MLC   & 3.00  & \textbf{2.17}  & 2.50  & 2.33  &  \\
    HMLC  & \textbf{2.00}  & 2.83  & 2.50  & 2.67  &  \\
    STR   & 3.00  & 3.33  & \textbf{1.83}  & 4.17  & 2.67 \\
    MTR   & 2.57  & 2.57  & \textbf{1.71}  & 3.14  &  \\
    \hline
    \end{tabular}%
\end{table}%
As explained in Sec.~\ref{sec:best-ens}, for the ensemble-based rankings, RFs are always preferable in terms of speed.
They can still be outperformed by Relief if the number of features is higher and the number of examples is moderate, which follows directly
from the $\mathcal{O}$-values in Secs.~\ref{sec:times-ensemble} and \ref{sec:times-relief}. All these methods are implemented in the Clus system (Java), whereas our implementation of the Laplace score is, as mentioned before, Python-based (Scikit Learn and numpy).
Thus, even though Laplace and Relief have the same core operations (finding nearest neighbors), using higly-optimized Scikit Learn's methods (such as $k$NN) puts
Laplace at the first place, whereas Relief is (second but) last, for STR problems.
\begin{table}[ht]
  \centering
  \caption{The average ranks of different SSL-ranking algorithms in terms of their induction times.
           Since the time complexity of ensemble-based rankings (almost) equals the induction time of the ensembles,
           we report the latter. For each task, we show the ranks for both extreme values of $L$.}
    \begin{tabular}{|c|r|rrrrr|}
    \hline
    task  & \multicolumn{1}{c|}{$L$} & \multicolumn{1}{l}{RFs} & \multicolumn{1}{l}{ETs} & \multicolumn{1}{l}{bagging} & \multicolumn{1}{l}{Relief} & \multicolumn{1}{l|}{Laplace} \\
    \hline
    \multirow{2}[2]{*}{classification} & 50    & \textbf{1.15}  & 2.46  & 3.15  & 3.23  &  \\
          & 500   &\textbf{ 1.31}  & 2.69  & 3.54  & 2.46  &  \\
    \hline
    \multirow{2}[2]{*}{MLC} & 50    & 1.67  & 3.50  & 3.33  & \textbf{1.50}  &  \\
          & 500   & 2.00  & 3.00  & 3.67  & \textbf{1.33}  &  \\
    \hline
    \multirow{2}[2]{*}{HMLC} & 50    & \textbf{1.17}  & 2.83  & 3.17  & 2.83  &  \\
          & 500   & \textbf{1.33}  & 3.00  & 3.50  & 2.17  &  \\
    \hline
    \multirow{2}[2]{*}{STR} & 50    & 2.17  & 3.50  & 3.83  & 4.50  & \textbf{1.00} \\
          & 500   & 2.67  & 3.17  & 4.67  & 3.50  & \textbf{1.00} \\
    \hline
    \multirow{2}[2]{*}{MTR} & 50    & \textbf{1.86}  & 2.29  & 3.71  & 2.14  &  \\
          & 500   & \textbf{1.71}  & 2.43  & 3.71  & 2.14  &  \\
    \hline
    \end{tabular}%
  \label{tab:times:all}%
\end{table}%

\section{Conclusions}\label{sec:conclusions}

In this work, we focus on \textbf{semi-supervised learning of feature ranking}. The feature rankings are learned in the context of simple (single-target) classification and regression as well as in the context of structured output prediction (multi-label classification, hierarchical multi-label classification and multi-target regression). This is the first work that treats the task of feature ranking within the semi-supervised structured output prediction - it treats all the different prediction tasks in an unified way.  

We propose, develop and evaluate \textbf{two approaches for SSL feature ranking for SOP} based on tree ensembles and the Relief family of algorithms. The tree ensemble-based rankings can be learned using three ensemble learning methods (Bagging, Random Forests, Extra Trees) coupled with three scoring functions (Genie3, Symbolic and random forest scoring). The Relief-based rankings use the regression variant of the Relief algorithm for extension towards the SOP tasks. This is the first extension of a Relief algorithm towards semi-supervised learning. 

An \textbf{experimental evaluation of the proposed methods is carried out on 38 benchmark datasets} from the five machine learning tasks: 13 from classification, 6 from multi-label classification, 6 from hierarchical multi-label classification, 6 from regression and 7 from multi-target regression. Whenever available, we compared the performance of the proposed methods to the performance of state-of-the-art methods. Furthermore, we compared the performance of the semi-supervised feature ranking methods with their supervised counterparts. 

The results from the extensive evaluation are best summarized through the answers of the research questions:
\begin{enumerate}
\item {\emph{For a given ensemble-based feature ranking score, which ensemble method is the most appropriate?}\\
Generally, Random Forests perform the best for the classification-like tasks (classification, muilti-label classification and hierarchical multi-label classification), while for the regression-like tasks (regression, multi-target regression) Extra-PCTs perform the best. Furthermore, across all tasks, Random Forests are the most efficient method considering induction times.}
\item {\emph{Are there any qualitative differences between the semi-supervised and supervised feature rankings?}\\
The semi-supervised rankings tend to capture a more global picture of the data, whereas the supervised ones reflect a more local one.}
\item {\emph{Can the use of unlabeled data improve feature ranking?}\\
Semi-supervised feature rankings outperform their supervised counterpart across a majority of the datasets from the different tasks.}
\item {\emph{Which feature ranking algorithm performs best?}\\
Different SSL feature ranking methods perform the best for the different tasks: Symbolic ranking is the best for the regression and multi-target regression, Random forest ranking for multi-label classification, Genie3 for hierarchical multi-label classification, and Relief for classification.}
\end{enumerate}


\begin{acknowledgements}
The computational experiments presented here were executed on a computing infrastructure from the Slovenian Grid (SLING) initiative, and we thank the administrators Barbara Kra\v{s}ovec and Janez Srakar for their assistance.
\end{acknowledgements}

%
%

\bibliographystyle{apalike}
\bibliography{references}   

\end{document}